\documentclass[10pt,twocolumn,letterpaper]{article}
\usepackage{wacv}

\def\wacvPaperID{582} 
\wacvfinalcopy
\ifwacvfinal
\fi

\usepackage{times}
\usepackage{epsfig}
\usepackage{graphicx}
\usepackage{amsmath}
\usepackage{amssymb}
\usepackage[pagebackref=true,breaklinks=true,letterpaper=true,colorlinks,bookmarks=false]{hyperref}

\usepackage[utf8]{inputenc}
\usepackage[T1]{fontenc}
\usepackage{hyperref} 
\usepackage{url}      
\usepackage{booktabs} 
\usepackage{amsfonts} 
\usepackage{nicefrac} 
\usepackage{microtype}
\usepackage{bm}
\usepackage{mathrsfs}
\usepackage{mathtools}
\usepackage{dsfont}
\usepackage{enumerate}
\usepackage[,slantedGreek]{mathptmx}
\usepackage{xspace}
\usepackage{mdframed}
\usepackage{verbatim}
\usepackage[ruled,vlined]{algorithm2e}
\usepackage[justification=centering]{subfig}

\DeclareMathOperator*{\argmin}{arg\,min}
\newtheorem{lemma}{Lemma}
\newtheorem{theorem}{Theorem}

\usepackage[font=small,labelfont=bf,tableposition=top]{caption}
\usepackage{babel}
\DeclareCaptionLabelFormat{andtable}{#1~#2  \&  \tablename~\thetable}

\makeatletter
\DeclareRobustCommand\onedot{\futurelet\@let@token\@onedot}
\def\@onedot{\ifx\@let@token.\else.\null\fi\xspace}

\makeatother

\def\wacvPaperID{582} 
\wacvfinalcopy

\ifwacvfinal
\fi

\ifwacvfinal
\usepackage[breaklinks=true,bookmarks=false]{hyperref}
\else
\usepackage[pagebackref=true,breaklinks=true,colorlinks,bookmarks=false]{hyperref}
\fi

\ifwacvfinal

\begin{document}

\title{Latent reweighting, an almost free improvement for GANs}

\author{Thibaut Issenhuth\textsuperscript{1,2}, Ugo Tanielian\textsuperscript{1}, David Picard \textsuperscript{2}, Jérémie Mary\textsuperscript{1}\\
\textsuperscript{1} Criteo AI Lab, Paris, France.\\
\textsuperscript{2} LIGM, Ecole des Ponts, Marne-la-Vallée, France. \\
{\tt\small \{thibaut.issenhuth,ugo.tanielian,jeremie.mary\}@criteo.com,david.picard@enpc.fr}
}
\maketitle
\ifwacvfinal\thispagestyle{empty}\fi

\begin{abstract}
Standard formulations of GANs, where a continuous function deforms a connected latent space, have been shown to be misspecified when fitting different classes of images. In particular, the generator will necessarily sample some low-quality images in between the classes. Rather than modifying the architecture, a line of works aims at improving the sampling quality from pre-trained generators at the expense of increased computational cost. 
Building on this, we introduce an additional network to predict latent importance weights and two associated sampling methods to avoid the poorest samples. This idea has several advantages: 1) it provides a way to inject disconnectedness into any GAN architecture, 2) since the rejection happens in the latent space, it avoids going through both the generator and the discriminator, saving computation time, 3) this importance weights formulation provides a principled way to reduce the Wasserstein's distance to the target distribution. We demonstrate the effectiveness of our method on several datasets, both synthetic and high-dimensional.
\end{abstract}

\section{Introduction}
GANs \cite{GANs} are an effective way to learn complex and high-dimensional distributions, leading to state-of-the-art models for image synthesis in both unconditional \cite{karras2018style} and conditional settings \cite{brock2018large}. However, it is well-known that a single generator with an unimodal latent variable cannot recover a distribution composed of disconnected sub-manifolds \cite{khayatkhoei2018disconnected}. This leads to a common problem for practitioners: the existence of very low-quality samples when covering different modes. This is formalized by \cite{tanielian2020learning} which refers to this area as the no GAN's land and provides impossibility theorems on the learning of disconnected manifolds with standard formulations of GANs. Fitting a disconnected target distribution requires an additional mechanism inserting disconnectedness in the modeled distribution. A first solution is to add some expressivity to the model: \cite{khayatkhoei2018disconnected} propose to  train a mixture of generators, while \cite{gurumurthy2017deligan} make use of a multi-modal latent distribution. 

\begin{figure}
    \centering
    \includegraphics[width=\columnwidth]{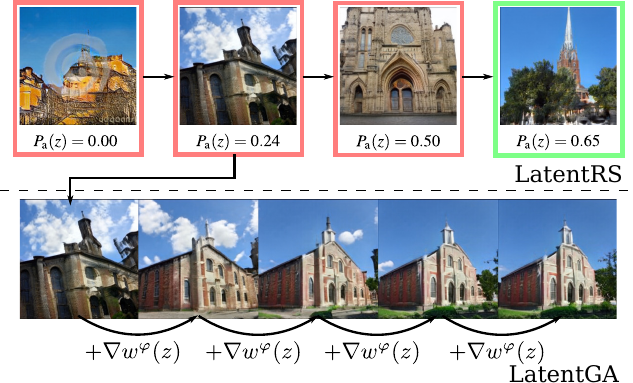}
    \caption{\textbf{Overview of the proposed method}. GANs tend to produce poor images for unlucky draws of the latent variable (top row, left). We introduce importance weights $w^\varphi(z)$ in the latent space that allow us to use rejection sampling and accept a given latent variable $z$ with probability $P_a(z) \propto w^\varphi(z)$ (LatentRS, top row), or to perform a simple gradient ascent over the importance weight (LatentGA, bottom row), leading to better images. Both strategies can be combined for improved image quality. Images generated with StyleGAN2 trained on LSUN Church.}
    \label{fig:example_intro}
\end{figure}

A second line of research relies heavily on a variety of Monte-Carlo algorithms, such as Rejection Sampling \cite{azadi2018discriminator} or Metropolis-Hastings  \cite{turner2019metropolis}. Monte-Carlo methods aim at sampling from a target distribution, while having only access to samples generated from a proposal distribution. Using the previously learned generative distribution as a proposal distribution, this idea was successfully applied to GANs. However, one of the main drawbacks is that Monte-Carlo algorithms only guarantee to sample from the target distribution under strong assumptions. First, we need access to the density ratios between the proposal and target distributions or equivalently to a perfect discriminator \cite{azadi2018discriminator}. Second, these methods are efficient only if the support of the proposal distribution fully covers the one of the target distribution. This is unlikely to be the case when dealing with high-dimensional datasets \cite{arjovsky2017towards}. 

To tackle this issue, we propose a novel method aiming at reducing the Wasserstein distance between the previously trained generative model and the target distribution. This is done via the adversarial training of a third network that learns importance weights in the latent space. Note that this network does not aim at increasing the support of the proposal distribution but at re-weighting the latent distribution, under a Wasserstein criterion. Thus, these importance weights define a new distribution in the latent space, from which we propose to sample with two complementary methods: latent rejection sampling (latentRS) and latent gradient ascent (latentGA). To better understand our approach, we illustrate its efficiency with simple examples. On the top of the Figure \ref{fig:example_intro}, we show samples coming from a pre-trained StyleGAN2 \cite{karras2018style} and their respective acceptance probability (latentRS). At the bottom, we exhibit a sequence of generated images while following a gradient ascent on the learned importance weights (latentGA).

Our contributions are the following:
\begin{itemize}
    \item We propose a novel approach that trains a neural network to directly modify the latent space of a GAN. This provides a principled way to reduce the Wasserstein distance to the target distribution. 
    \item We show how to sample from this new generative model with different methods: latent Rejection Sampling (latentRS), latent Gradient Ascent (latentGA), and latentRS+GA, a method that leverages the complementarity between the two previous solutions.
    \item  We run a large empirical comparison between our proposed methods and previous approaches on a variety of datasets and distributions. We empirically show that all of our proposed solutions significantly reduce the computational cost of inference. More interestingly, our solutions propose a wide span of performances ranging from latentRS, optimizing speed, that matches state-of-the-art almost for free (computational cost divided by 15) and latentRS+GA (computational cost divided by 3) that outperforms previous approaches.
\end{itemize}

\textbf{Notation.} Before moving to the related work section, we shortly present the notation needed in the paper. The goal of the generator is to generate data points that are ``similar'' to samples collected from some target probability measure $\mu_\star$. The measure $\mu_\star$ is defined on a potentially high-dimensional space $\mathds{R}^D$, equipped with the euclidean norm $\|\cdot\|$. We call $\mu_n$ the empirical measure. To approach $\mu_\star$, we use a parametric family of generative distribution, where each distribution is the push-forward measure of a latent distribution $Z$ and a continuous function modeled by a neural network. 
In most applications, the random variable $Z$ defined on a low-dimensional space $\mathds{R}^d$ is either a multivariate Gaussian distribution or uniform distribution. The generator is a parameterized class of functions from $\mathds{R}^d$ to $\mathds{R}^D$, say $\mathscr{G}= \{G_\theta: \theta \in \Theta \}$, where $\Theta \subseteq \mathds{R}^p$ is the set of parameters describing the model. Each function $G_\theta$ takes input from $Z$ and outputs ``fake'' observations with distribution $\mu_\theta = G_\theta \sharp Z$. On the other hand, the discriminator is described by a family of functions from $\mathds{R}^D$ to $\mathds{R}$, say $\mathscr{D} = \{D_\alpha: \alpha \in \Lambda \}$, $\Lambda \subseteq \mathds{R}^Q$. 
Finally, for any given distribution $\mu$, we note $S_\mu$ its support. 

\section{Related Work}
\cite{GANs} already stated that when training vanilla GANs, the generator could ignore modes of the target distribution: this is called mode collapse. A significant step towards understanding this phenomenon was made by \cite{arjovsky2017towards} who explained that the standard formulation of GANs leads to vanishing or unstable gradients. The authors proposed the Wasserstein GANs (WGANs) architecture \cite{arjovsky2017wasserstein} where, in particular, discriminative functions are restricted to the class of 1-Lipschitz functions. WGANs aim at solving the following:
\begin{equation}\label{eq:WGANs}
    \underset{\alpha \in A}{\sup} \ \underset{\theta \in \Theta}{\inf} \ \mathds{E}_{x \sim \mu_\star} \ D_\alpha(x) - \mathds{E}_{z \sim \gamma} \ D_\alpha(G_\theta(z))
\end{equation}

\subsection{Learning disconnected manifolds with GANs: training and evaluation}
The broader drawback of standard GANs is that, since any modeled distribution is the push-forward of a unimodal distribution by a continuous transformation, it has a connected support. This means that when the generator covers multiple disconnected modes of the target distribution, it necessarily generates samples out of the real data manifold \cite{khayatkhoei2018disconnected}. Consequently, any thorough evaluation of GANs should assess simultaneously both the quality and the variety of the generated samples. 
To solve this issue, \cite{sajjadi2018assessing} and \cite{kynkaanniemi2019improved} propose a Precision/Recall metric that aims at measuring both the \textit{mode dropping} and the \textit{mode inventing}. The precision refers to the portion of generated points that belongs to the target manifold, while the recall measures how much of the target distribution can be reconstructed by the model distribution. 

Building on this metric, \cite{tanielian2020learning} highlighted the trade-off property of GANs deriving upper-bounds on the precision of standard GANs. To solve this problem, a common direction of research consists in over-parameterizing the generative model. \cite{khayatkhoei2018disconnected} enforces diversity by using a mixture of generators, while \cite{gurumurthy2017deligan} suggests that a mixture of Gaussians in the latent space is efficient to learn diverse and limited data. Similarly, \cite{balaji2020robust} propose importance weights that aim at robustifying the training of GANs and make it less sensitive to the target distribution's outliers.

\subsection{Improving the quality of GANs post-training} 
Another line of research consists in improving the sampling quality of pre-trained GANs. \cite{tanielian2020learning} proposed a heuristic to insert disconnectedness and remove the samples mapped out of the true manifold. \cite{tanaka2019discriminator} designed Discriminator Optimal Transport (DOT), a gradient ascent driven by a Wasserstein discriminator to improve every single sample. Similarly, \cite{che2020your} follow a discriminator-driven Langevin dynamic.

Another well-studied possibility would be to use Monte-Carlo (MC) methods \cite{robert2013monte}. Following this path, \cite{azadi2018discriminator} were the first to use a rejection sampling method to improve the quality of the proposal distribution $\mu_\theta$. The authors use the fact that the optimal vanilla discriminator trained with binary cross-entropy is equal to $\mu_\star/(\mu_\star + \mu_\theta)$. Thus, a parametric discriminator $D_\alpha: \mathbb{R}^D \to [0,1]$ can be used to approximate the density ratios $r_\alpha$ as follows:
\begin{equation}
    r_\alpha(x) := \frac{\mu_\star(x)}{\mu_\theta(x)} = \frac{D_\alpha(x)}{1-D_\alpha(x)}.
\end{equation}
This density ratio can then be plugged in the Rejection Sampling (RS) algorithm. Doing so, it can be shown that sampling from $\mu_\theta$ and accepting samples probabilistically is equivalent to sample from the target distribution $\mu_\star$. The acceptance probability of a given sample $x$ is $\mathds{P}_a(x)= \frac{r_\alpha(x)}{k}$. This is valid as long as there is a constant $k \in \mathds{R}^\texttt{+}$ such that $\mu_\star(x) \leq k \mu_\theta(x)$ for all x. 

\cite{turner2019metropolis} use similar density ratios and derive MH-GAN, by using the independent Metropolis-Hasting algorithm \cite{hastings1970monte}. Finally, \cite{grover2019bias} use these density ratios $r_\alpha$ as importance weights and perform discrete sampling relying on the Sampling-Importance-Resampling (SIR) algorithm \cite{rubin1988using}. Given $X_1, \hdots, X_n \sim \mu_\theta^n$, we have:
\begin{equation}
    \mu_{\theta, \alpha}^{\text{SIR}}(X_i) = \frac{r_\alpha(X_i)}{\sum\limits_{j=1}^n r_\alpha(X_j)}.
\end{equation}
Note that these algorithms all rely on similar density ratios and differ by the acceptance-rejection scheme chosen. Interestingly, in RS, the acceptance rate is not controlled, but we are guaranteed to sample from $\mu_\star$. Conversely, with SIR and MH, the acceptance rate is a chosen parameter, but we are sampling from an approximation of the target distribution. 

\subsection{Drawbacks of density-ratio-based methods \label{drawbacks_density}}
Even though these methods have the advantage of being straightforward, they suffer from one main drawback. In practice, because both the target and the proposal manifold do not have full dimension in $\mathds{R}^D$ \cite{fefferman2016testing}, \cite[Lemma 3]{arjovsky2017towards} show that it is highly likely that $\mu_\theta(S_{\mu_\theta} \bigcap S_{\mu_\star})=0$ and $\mu_\star(S_{\mu_\theta} \bigcap S_{\mu_\star})=0$. Consequently, when dealing with high-dimensional datasets, the proposal distribution $\mu_\theta$ and the target distribution $\mu_\star$ might intersect on a null set. Thus, one would have $r_\alpha(x) = 0$ almost everywhere on $S_{\mu_\theta}$. In this setting, the assumptions of MC methods are broken, and these algorithms will not allow sampling from $\mu_\star$.

In order to correct this drawback, our method proposes to avoid the computation of density ratios from a classifier and to directly learn how to re-weight the proposal distribution. Our proposed scheme aims at minimizing the Wasserstein distance to the empirical measure while controlling the range of these importance weights.

\section{Adversarial Learning of \\
Latent Importance weights}
Similar to previous works, our method aims at improving the performance of a generative model, post-training. We assume the existence of a WGAN model $(G_\theta, D_\alpha)$ pre-trained using \eqref{eq:WGANs}. The pushforward generative distribution $\mu_\theta$ is assumed to be an imperfect approximation of the target distribution. The goal is now to learn how to redistribute the mass of the modeled distribution so that it best fits the target distribution. 

\subsection{Definition of the method}
To improve the sampling quality of our pre-trained GANs, we propose to learn an importance weight function that directly learns how to avoid low-quality images and focus on very realistic ones. More formally, we over-parameterize the class of generative distributions and define a parametric class $\Omega = \{w^\varphi, \varphi \in \Phi\}$ of importance weight functions. Each function $w^\varphi$ associates importance weights to latent space variables and is defined from $\mathds{R}^d$ to $\mathds{R}^\texttt{+}$. For a given latent space distribution $\gamma$ and a network $w^\varphi$, a new measure $\gamma^\varphi$ is defined on $\mathds{R}^d$:
\begin{equation}\label{eq:gamma_varphi}
    \text{for all $z \in \mathds{R}^d$, } \rm{d} \gamma^\varphi(z) = w^\varphi(z) \rm{d} \gamma(z)
\end{equation}
Using this formulation, we can prove the following lemma:
\begin{lemma}\label{lem:lemma1}
    Assume that $\mathds{E}_{\gamma} \ w^\varphi = 1$, then the measure $\gamma^\varphi$ is a probability distribution defined on $\mathds{R}^d$.
\end{lemma}
Consequently, we now propose a new modeled generative distribution $\mu_\theta^\varphi$, the pushforward distribution $\mu_\theta^\varphi = G_\theta \sharp \gamma^\varphi$. The objective is to find the optimal importance weights $w^\varphi$ that minimizes the Wasserstein distance between the true distribution $\mu_\star$ and the new class of generative distributions. The proposed method can thus be seen as minimizing the Wasserstein distance to the target distribution, over an increased class of generative distributions. Denoting by $\text{Lip}_1$ the set of $1$-Lipschitz real-valued functions on $\mathds{R}^D$, i.e.,
\begin{equation*}
    \text{Lip}_1 = \big\{f: \mathds{R}^D \to \mathds R: \frac{|f(x)-f(y)|}{\|x-y\|} \leqslant 1, \ (x \neq y) \in (\mathds{R}^D)^2\big\},
\end{equation*}
we want, given a pre-trained model $\mu_\theta$, to solve:
\begin{align*}
    \underset{\varphi \in \Phi}{\argmin} \ W(\mu_\star, \mu_\theta^\varphi)  &= \underset{w^\varphi \in \Omega}{\argmin} \ \underset{D \in \text{Lip}_1}{\sup} \ \mathds{E}_{\mu_\star} D  - \mathds{E}_{\mu_\theta^\varphi} D \\
    &= \underset{w^\varphi \in \Omega}{\argmin} \ \underset{D \in \text{Lip}_1}{\sup} \  \mathds{E}_{\mu_\star} D - \mathds{E}_{\mu_\theta} w^\varphi D
\end{align*}
The network $w^\varphi$, parameterized using a feed-forward neural network, thus learns how to redistribute the mass of $\mu_\theta$ such that $\mu_\theta^\varphi$ is closer to $\mu_\star$ in terms of Wasserstein distance. Similarly to the WGANs training, the discriminator $D_\alpha$ approximates the Wasserstein distance. $D_\alpha$ and $w^\varphi$ are trained adversarially, whilst keeping the weights of $G_\theta$ frozen, using the following optimization scheme:
\begin{equation}\label{eq:w_phi}
    \underset{\varphi \in \Phi}{\inf} \ \underset{\alpha \in \Lambda}{\sup} \ \mathds{E}_{x \sim \mu_\star} D_\alpha(x) - \mathds{E}_{z \sim Z} \ w^\varphi(z) \times D_\alpha(G_\theta(z))
\end{equation}
Note that our formulation can also be plugged on top of any objective function used for GANs. 

\subsection{Optimization procedure}
However, as in the field of counterfactual estimation, a naive optimization of importance weights by gradient descent can lead to trivial solutions.
\begin{enumerate}
    \item First, if for example, the Wasserstein critic $D_\alpha$ outputs negative values for any generated sample, the network $w^\varphi$ could simply learn to avoid the dataset and output $0$ everywhere \cite{swaminathan2015batch}.
    \item Second, another problem comes from the fact that \eqref{eq:w_phi} can be minimized not only by putting large importance weights $w^\varphi(z)$ on the examples with high likelihoods $D_\alpha(G(z))$ but also by maximizing the sum of the weights: this is the propensity overfitting \cite{Swaminathan2015self}. 
    \item For the objective defined in \eqref{eq:w_phi} to be a valid Wasserstein distance minimization scheme, the measure $\mu_\theta^\varphi$ must be a probability distribution, \textit{i.e.} $\mathds{E}_\gamma w^\varphi=1$.
\end{enumerate}
To tackle this, we first add a penalty term in the loss to enforce the expectation of the importance weights to be close to $1$. This is similar to the self-normalization proposed by \cite{Swaminathan2015self}. However, one still has to cope with the setting where the distribution $\gamma^\varphi$ collapses to discrete data points: 
\begin{theorem}\label{th:theorem1}
    Given a pre-trained generative distribution $\mu_\theta$ absolutely continuous with respect to the Lebesgue measure on $\mathds{R}^D$. Let $\Phi$ be the non-parametric class of continuous functions satisfying $\mathds{E}_\gamma w^\varphi=1$. We have that:
    \begin{equation*}
        W(\mu_n, \frac{1}{n} \sum_{i=1}^n \delta(\Tilde{X_i})) \leqslant \underset{\varphi \in \Phi}{\inf} W(\mu_n, \mu_\theta^\varphi)
    \end{equation*}
    where $\delta$ refers to the Dirac probability distribution and $\Tilde{X_i} = \underset{x \in S_{\mu_\theta}}{\argmin} \ \|x-X_i\|$.
\end{theorem}
For clarity, the proof is delayed in Appendix. Intuitively, this theorem shows that the best way to approximate the empirical measure $\mu_n$ would be by considering a mixture of Diracs with each mode being the projection of a training data point on the support of the learned manifold $S_{\mu_\theta}$. The network $w^\varphi$ could thus be tempted to approximate this mixture of Diracs defined in Theorem \ref{th:theorem1} and collapse on some specific latent data points. This could lead to an increased time complexity at inference (see \cite[Section 3]{azadi2018discriminator}). More importantly, this would mean \textit{a mode collapse} and a lack of diversity in the generated samples. 

To avoid such cases where small areas of $z$ have really high $w^\varphi(z)$ values (mode collapse), we enforce a soft-clipping on the weights \cite{bottou2013counterfactual, grover2019bias}. Note that this constraint on $w^\varphi(z)$ could also be implemented with a bounded activation function on the final layer, such as a re-scaled sigmoid or tanh activation. Finally, we get the following objective function for the network $w^\varphi$:
\begin{align}\label{eq:full_training_w_varphi_and_regul}
    \underset{\varphi \in \Phi}{\sup} \ \mathds{E}_{z \sim Z} \ & \underbrace{ w^\varphi(z) \big(D_\alpha(G_\theta(z)) - \Delta \big)}_{\text{discriminator reward}} - \lambda_1 \underbrace{ \big(\mathds{E}_{z \sim Z} w^\varphi(z) - 1 \big)^2}_{\text{self-normalization}} \nonumber \\
    & - \lambda_2 \underbrace{\mathds{E}_{z \sim Z} \max \big(0, (w^\varphi(z) - m) \big)^2}_{\text{soft-clipping}}, 
\end{align}
where $\Delta = \min_{z \sim Z} \ D_\alpha(G(z))$. $\lambda_1$, $\lambda_2$, and $m$ are hyper-parameters (values displayed in Appendix). For more details, we refer the reader to Algorithm \ref{algo:training}.

\begin{algorithm}
\SetAlgoLined
 \textbf{Require:} Data $\mu_n$, Prior $Z$,  Gen. $G_\theta$, Disc. $D_\alpha$, number of $D_\alpha$ updates $n_d$, soft-clipping param. $m$, regularization weights $\lambda_1$ and $\lambda_2$, batch size $b$\;
 \While{$\varphi$ has not converged}{
  \For{ $i=0,...,n_d$}{
  Sample real data $\{x_i\}_{i=1}^b \sim \mu_n$\; 
  Sample latent vectors $\{z_i\}_{i=1}^b \sim Z$ \; 
   $\text{EMD} \gets \frac{1}{b} \sum_{i=1}^b D_\alpha(x_i) -  w^\varphi(z_i) D_\alpha(G_\theta(z_i)) $\; 
   $\text{GP} \gets \text{Gradient-Penalty}(D_\alpha,x,G_\theta(z))$\;
   $\text{grad}_\alpha \gets \nabla_\alpha (- \text{EMD} + \text{GP}) $ \;
   $\text{Update } \alpha \text{ with grad}_\alpha$;
  }
  Sample $\{z_i\}_{i=1}^b \sim Z$ \; 
   $\Delta \gets \text{min}_i [D_\alpha(G_\theta(z_i))]$ \;
   $\text{EMD} \gets \frac{1}{b} \sum_{i=1}^b w(z_i) [D_\alpha(G_\theta(z_i)) - \Delta]$\;
   $R_{norm} \gets ([\frac{1}{b}\sum_{i=1}^b w(z_i)] - 1)^2$ \;
   $R_{clip} \gets \frac{1}{b}\sum_{i=1}^b \text{max}(0,w^\varphi(z_i)- m)^2$ \;
   $\text{grad}_\varphi \gets \nabla_\varphi ( \text{EMD} + \lambda_1 R_{norm} + \lambda_2 R_{clip})$ \;
   $\text{Update } \varphi \text{ with grad}_\varphi$; 
 }
\caption{Adversarial learning of $w^\varphi$ \label{algo:training}}
\end{algorithm}

\subsection{Sampling from the latent importance weights}
Given a pre-trained generator $G_\theta$ and an importance network $w^\varphi$, we now present the three proposed sampling algorithms associated with our model: 

\paragraph{1) Latent Rejection Sampling (latentRS, Algorithm \ref{algo:lrs}).} The first proposed method aims at sampling from the newly learned latent distribution $\gamma^\varphi$ defined in \eqref{eq:gamma_varphi}. Since the learned importance weights are capped by $m$ defined in \eqref{eq:full_training_w_varphi_and_regul}, this setting fits in the Rejection Sampling (RS) algorithm \cite{robert2013monte}. Any sample $z \sim \gamma$ is now accepted with probability $\mathds{P}_{\text{a}}(z) = w^\varphi(z)/m$. Interestingly, by actively capping the importance weights as it is done in counterfactual estimation \cite{bottou2013counterfactual, faury2019distributionally}, one controls the acceptance rates $\mathds{P}_a(z)$ of the rejection sampling algorithm:
\begin{equation*}
    \mathds{E}_\gamma \ \mathds{P}_{\text{a}}(z) = \int_{\mathds{R}^d} \frac{w^\varphi(z)}{m} \rm d\gamma(z) = \frac{1}{m}.
\end{equation*}

\begin{algorithm}
\SetAlgoLined
 \textbf{Requires:} Prior Z, Gen. $G_\theta$, Importance weight network $w^\varphi$, maximum importance weight $m$\;
 \While{True}{
  Sample $z \sim Z$ \;
  Sample $\alpha \sim \text{Uniform}[0,1]$ \;
  \If{$\frac{w^\varphi(z)}{m} \geq \alpha$}{
   break\;
   }
 }
 $x \gets G_\theta(z)$\;
 \KwResult{Selected point x}
 \caption{LatentRS \label{algo:lrs}}
\end{algorithm}

\paragraph{2) Latent Gradient Ascent (latentGA).} Inspired from \cite[Algorithm 2]{tanaka2019discriminator}, we propose a second method, latentGA, where we perform gradient ascent in the latent space (see the algorithm in Appendix). For any given sample in the latent space, we follow the path maximizing the learned importance weights. This method is denoted latentGA. Note that the learning rate and the number of updates used for this method are hyper-parameters that need to be tuned.  

\paragraph{3) Combining latentRS with Gradient Ascent (latent RS+GA, see Appendix).} Finally, we propose to combine sequentially both methods. In a first step, we avoid low-quality samples with latentRS. Then, we use latentGA to further improve the remaining generated samples. See algorithm in Appendix. 

\subsection{Advantages of the proposed approach}
We now discuss two advantages of our method compared to previous density-ratio-based Monte-Carlo methods. 

\paragraph{Computational cost.} By using sampling algorithms in the latent space, we avoid going through both the generator and the discriminator, leading to a significant computational speed-up. This is of particular interest when dealing with high-dimensional spaces, since we do not need to pass through deep CNNs generator and discriminator \cite{karras2018style}. In the next experimental section, we observe a computational cost decreased by a factor of 10.

\begin{figure*}[h] 
\centering 
\subfloat[Synthetic WGAN: real samples in green and fake ones in blue. \label{fig:wgan}] 
    { \includegraphics[width=0.224\textwidth]{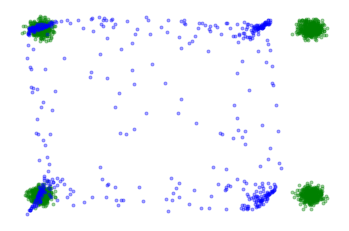} } 
\subfloat[MC method optimizing for a precision criterion \cite{azadi2018discriminator}. \label{fig:drs}] 
    { \includegraphics[width=0.224\textwidth]{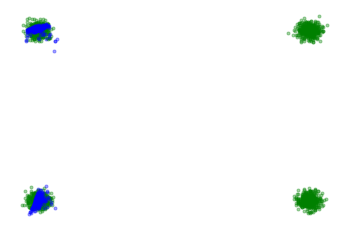} } 
\subfloat[Optimizing for Wasserstein criterion with latentRS (ours $\star$ ).\label{fig:latentrs}] 
    {     \includegraphics[width=0.224\textwidth]{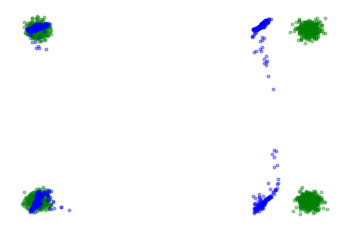} } 
\subfloat[Heatmap of the $w^\varphi$ in the latent space (in the blue areas, $w^\varphi$=0). \label{fig:latentweights}] 
    { \includegraphics[width=0.20\textwidth]{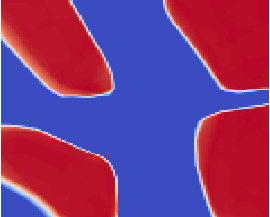} } 
\caption{Synthetic experiment mimicking the setting of GANs in high-dimension, where data and generated manifolds are close but do not perfectly intersect. While DRS only selects the intersection of manifolds and ignores the rest, the latent importance weights define a rejection mechanism that minimizes the Wasserstein distance. For conciseness, WGAN stands for WGAN-GP. \label{fig:figure1} } \end{figure*}

\paragraph{Monte-Carlo methods do not properly work when the support $S_{\mu_\theta}$ does not fully cover the support $S_{\mu_\star}$.} To better illustrate this claim, we consider a simple 2D motivational example where the real data lies on four disconnected manifolds. We start with a proposal distribution (in blue) that does not fully recover the target distribution (Figure \ref{fig:wgan}). In this setting, we see in Figure \ref{fig:drs} that the discriminator's density-ratio-based methods \cite{azadi2018discriminator} avoids half of the proposal distribution, while our proposed method learns a very different re-weighting (see Figure \ref{fig:latentrs}). 

This illustration is important since \cite[Theorem 2.2]{arjovsky2017towards} have shown that in high-dimension the intersection $S_{\mu_\star} \bigcap S_{\mu_\theta}$ is likely to be a negligible set under $\mu_\theta$. Knowing that $S_{\mu_\theta}$ does not fully recover $S_{\mu_\star}$, there is thus no theoretical guarantee that using a sampling algorithm will improve the estimation of $\mu_\star$. On the opposite, our method looks for the optimal re-weighting of $\mu_\theta$ under a well-defined criterion: the Wasserstein distance. This results in a better fit of the real data distribution (see next section). 

\section{Experiments}
In this section, we illustrate the efficiency of the proposed methods, latentRS, latentGA, and latentRS+GA on both synthetic and natural image datasets. On image generation tasks, we empirically stress that latentRS slightly surpasses density-ratio-based methods with respect to the Earth Mover's distance while reducing the time complexity by a factor of around 10. The use of latentGA also gives interesting experimental visualizations and improves image quality. More importantly, when combined, we show that latenRS+GA surpasses the concurrent methods, while still being less computationally intensive. Finally, we show results with different models such as Progressive GAN~\cite{karras2017progressive} and StyleGAN2~\cite{karras2020analyzing}.

\subsection{Evaluation metrics}
To measure the performances of GANs when dealing with low-dimensional applications, we equip our space with the standard Euclidean distance. However, for the case of image generation, we follow \cite{brock2018large, kynkaanniemi2019improved}  and consider the euclidean distance between embeddings of a pre-trained network, that convey more semantic information. Thus, for a pair of images $(a,b)$, we define the distance $d(a,b)$ as $d(a,b) = \| \phi(a) - \phi(b) \|_2$ where $\phi$ is a pre-softmax layer of a supervised classifier. On MNIST and F-MNIST, the classifier is pre-trained on the given dataset. On CelebA and LSUN Church, we use VGG-16 pre-trained on ImageNet.  

To begin with, we report the FID \cite{heusel2017gans}. We also compare the performance of the different methods with the Precision/Recall (PR) metric \cite{kynkaanniemi2019improved}. It is a more robust version of the Precision/Recall metric, which was first applied in the context of GANs by \cite{sajjadi2018assessing}. 
Finally, we approximate the Wasserstein distance using the Earth Mover's Distance (EMD) between generated and real data points. This measure is particularly suited to the study of WGANs, since it is linked to their objective function. Letting $X = \{x_1, \hdots, x_n\}$ and $Y= \{y_1, \hdots, y_n\}$ be two collections of $n$ data points and $\mathcal{S}$ be the set of permutations of $[1,n]$, the Earth Mover's distance between $X$ and $Y$ is defined by:
\begin{align*}
    \text{EMD}(X,Y) &= \underset{\sigma \in \mathcal{S}}{\min} \ \sum_{i=1}^n \|x_i-y_{\sigma_i}\|
\end{align*}

\subsection{Synthetic datasets}
\begin{table}
\begin{tabular}{l|c|c|}
  &  EMD & EMD 
\\ 
  &  Swiss Roll & 25 Gaussians \\
\hline 
WGAN  & $0.030 {\scriptstyle  \pm 0.002}$ & $0.044 {\scriptstyle  \pm 0.001}$ \\ \hline
WGAN: DRS  & $0.036 {\scriptstyle  \pm 0.004}$  & $0.038 {\scriptstyle  \pm 0.002}$\\
WGAN: SIR & $0.037 {\scriptstyle  \pm 0.003}$ & $0.041 {\scriptstyle  \pm 0.001}$\\
WGAN: DOT & $0.029 {\scriptstyle  \pm 0.003}$ & $\mathbf{0.035 {\scriptstyle  \pm 0.002}}$\\
WGAN: latentRS \small{($\star$)} & $\textbf{0.025} {\scriptstyle  \pm 0.002}$ & $0.036 {\scriptstyle  \pm 0.001}$\\
\hline
\end{tabular}
\caption{Comparison of latentRS with concurrent methods on two synthetic datasets in the same setting as DOT \cite{tanaka2019discriminator}. Our method enables a consistent gain in EMD, surpassing other methods on Swiss Roll and slightly behind DOT on Mixture of 25 Gaussians. For conciseness, WGAN stands for WGAN-GP. \label{table:synthetic_datasets}}
\end{table} 
To begin the experimental study, we test our method on 2D synthetic datasets in the same setting as \cite{tanaka2019discriminator}. Table \ref{table:synthetic_datasets} compares the latentRS method with previous approaches on the Swiss roll dataset and on a mixture of 25 Gaussians. We see that the network $w^\varphi$ efficiently redistributes the pre-trained distribution $\mu_\theta$ since $\text{EMD}(\mu_n, \mu_\theta^\varphi)$ is significantly smaller than $\text{EMD}(\mu_n, \mu_\theta)$. 

\subsection{Image datasets}

\begin{table*}
\begin{center}
\begin{tabular}{|l|c|c|c|c|c|}
\cline{2-6}
\multicolumn{1}{l|}{\textbf{CelebA 128x128}}  & Prec. ($\uparrow$) & Rec. ($\uparrow$) &  EMD ($\downarrow$) & FID ($\downarrow$) & Inference (ms) \\
\hline
ProGAN & $74.2 {\scriptstyle \pm 0.9}$ &  ${60.7 {\scriptstyle \pm 1.4}}$ & $25.4 {\scriptstyle \pm 0.1}$ & $11.30  {\scriptstyle \pm 0.02}$ & $3.6$\\ \hline
ProGAN: SIR & ${79.5 {\scriptstyle \pm 0.4}}$ &  $\mathbf{57.3 {\scriptstyle \pm 1.0}}$ & $24.9 {\scriptstyle \pm 0.2}$ & $12.01 {\scriptstyle \pm 0.04}$ & $49.0$ \\
ProGAN: DOT & ${81.3 {\scriptstyle \pm 1.0}}$ &  $52.9 {\scriptstyle \pm 1.4}$ & $25.0 {\scriptstyle \pm 0.1}$ & $11.01{\scriptstyle \pm 0.03}$ & $67.6$ \\
ProGAN: latentRS ($\star$) & $80.4 { {\scriptstyle \pm 0.9}}$ &  $55.7 {\scriptstyle \pm 1.0}$ & ${24.7 {\scriptstyle \pm 0.1}}$ & ${10.77 {\scriptstyle \pm 0.04}}$ & $\mathbf{4.5}$ \\ 
ProGAN: latentRS+GA ($\star$) & $\mathbf{83.3 {\scriptstyle \pm 1.0}}$ &  $52.7 {\scriptstyle \pm 0.9}$ & $\mathbf{24.5 {\scriptstyle \pm 0.1}}$ & $\mathbf{10.75 {\scriptstyle \pm 0.04}}$ & $20.5$ \\
\hline
\multicolumn{4}{l|}{\textbf{LSUN Church 256x256}}  \\
\hline
StyleGAN2 & $55.6 {\scriptstyle \pm 1.2}$ &  ${62.4 {\scriptstyle \pm 1.1}}$ & $23.6 {\scriptstyle \pm 0.1}$ & $6.91 {\scriptstyle \pm 0.02}$ & $11.7$ \\ \hline
StyleGAN2: SIR & $60.5 {\scriptstyle \pm 1.4}$ & $\mathbf{58.1 {\scriptstyle \pm 1.3}}$ & $23.4 {\scriptstyle \pm 0.1}$ & $7.36 {\scriptstyle \pm 0.01}$ & $130.0$\\ 
StyleGAN2: DOT & $67.4 {\scriptstyle \pm 1.4}$ & $48.3 {\scriptstyle \pm 1.0}$ & $23.1 {\scriptstyle \pm 0.1}$ & $6.85 {\scriptstyle \pm 0.02}$ & $196.7$ \\ 
StyleGAN2: latentRS ($\star$) & $63.3 {\scriptstyle \pm 0.7}$ & $57.7 {\scriptstyle \pm 1.0}$ & $23.1 {\scriptstyle \pm 0.1}$ & $6.31 {\scriptstyle \pm 0.02}$ & $\mathbf{16.2}$ \\ 
StyleGAN2: latentRS+GA ($\star$) & $\mathbf{72.6 {\scriptstyle \pm 1.1}}$ & $43.2 {\scriptstyle \pm 1.3}$ & $\mathbf{22.6 {\scriptstyle \pm 0.1}}$ & $\mathbf{6.27 {\scriptstyle \pm 0.03}}$ & $43.2$ \\ 
\hline
\end{tabular}
\end{center}
\caption{latentRS+GA is the best performer, and latentRS matches SOTA with a significantly reduced inference cost (by an order of at least 10). $\pm$ is $97\%$ confidence interval. Inference refers to the time in milliseconds needed to compute one image on a NVIDIA V100 GPU. \label{table:real_worlds_results_main}}
\end{table*}

\textbf{Implementation of baselines.} We now compare latentRS, latentGA, and latentRS+GA with previous works leveraging discriminator's information on high-dimensional data. In particular, we implemented a wide set of post-processing methods for GANs: DRS \cite{azadi2018discriminator}, MH-GAN \cite{turner2019metropolis}, SIR \cite{grover2019bias} and DOT \cite{tanaka2019discriminator}. DRS, MH-GAN and SIR use the same density ratios, and we did not see significant differences between those three methods in our experiments. Consequently, for the following experiments, we compare our algorithms to SIR and DOT. For SIR, we take the discriminator at the end of the adversarial training, fine-tune it with the binary cross-entropy loss and select the best model in terms of EMD. Overall, we explicitly follow the framework used by \cite{azadi2018discriminator,grover2019bias}: we keep the gradient penalty \cite{gulrajani2017improved},  spectral normalization \cite{spectral_normGANs} during fine-tuning and do not include an explicit mechanism to calibrate the classifier.
  
\begin{figure}
    \subfloat
    {   
        \includegraphics[width=0.5\linewidth]{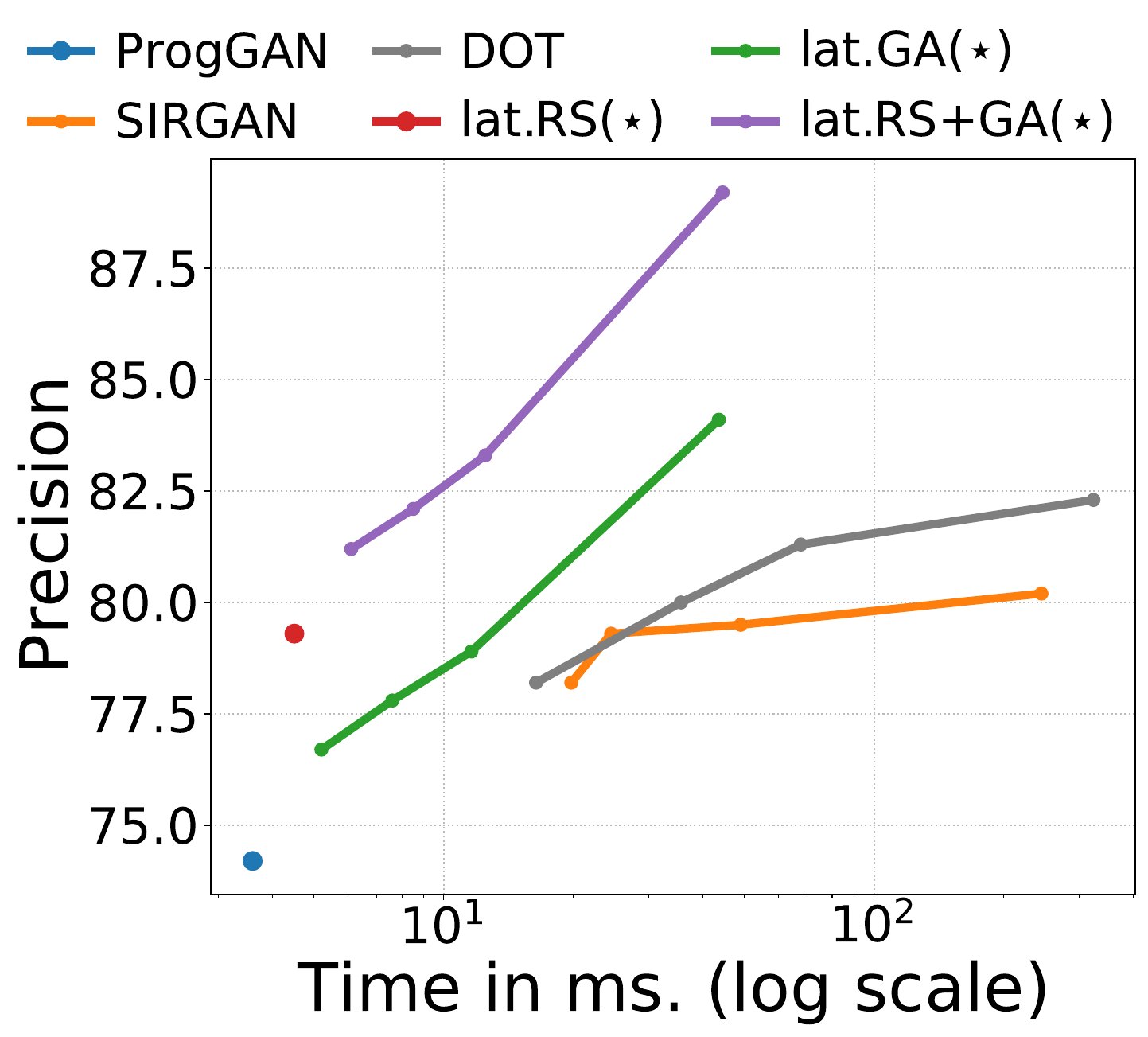}
    }
    \subfloat
    {   
        \includegraphics[width=0.5\linewidth]{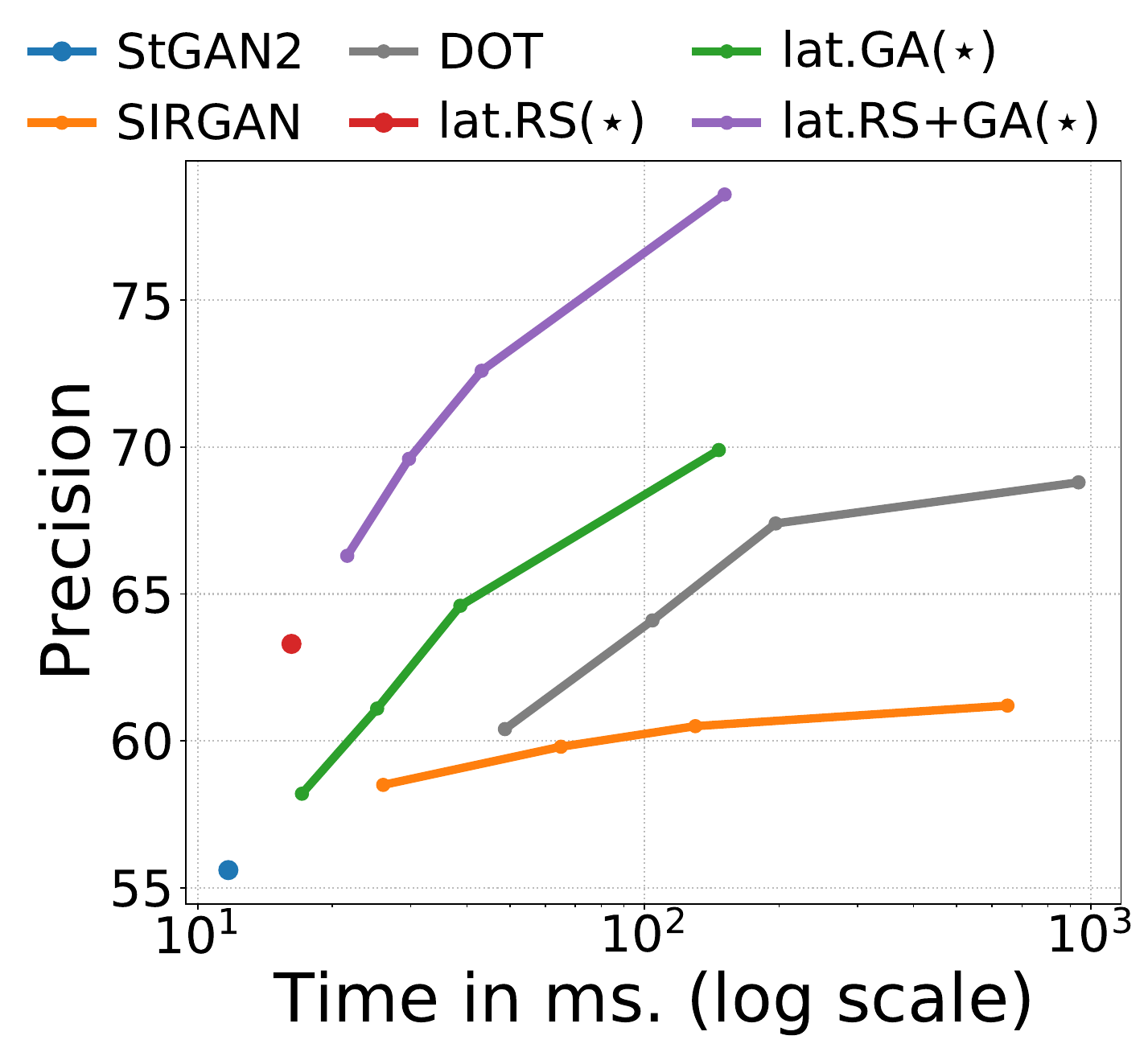}
    }
    \caption{Visualization of the trade-off between the time spent to generate an image and its average precision. Interestingly, latentRS+GA has the best Pareto front. Left: ProGAN trained on CelebA. Right: StyleGan2 trained on LSUN Church. \label{fig:comparison}}
\end{figure}

\captionsetup[subfigure]{labelformat=empty}
\begin{figure*}[h]
    \centering
    \subfloat[$P_a(z) = 0.00$]
    {   
        \includegraphics[width=0.11\linewidth,trim={0cm 1.55cm 0 0},clip]{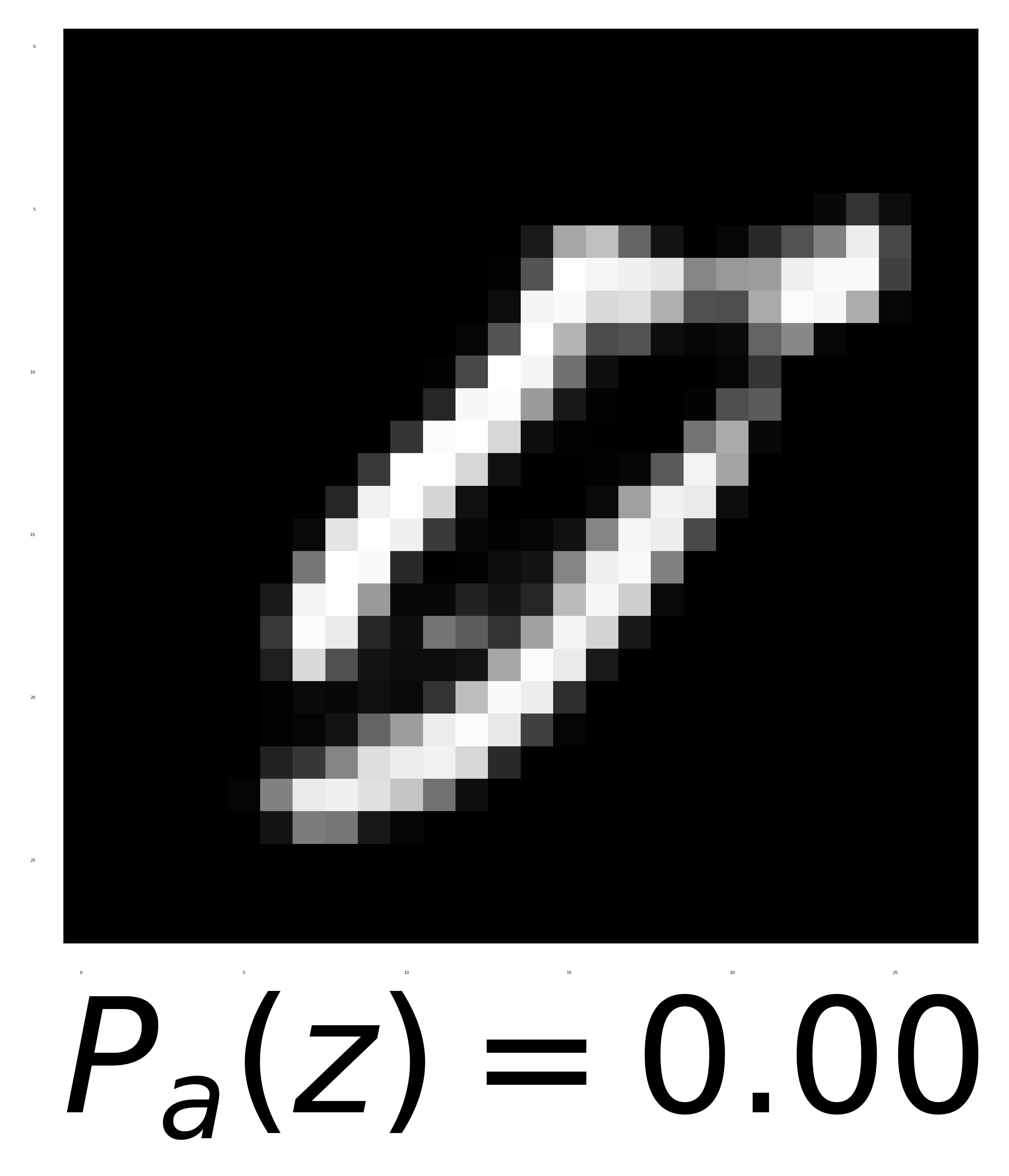}
    } 
    \subfloat[$P_a(z) = 0.05$]
    {   
        \includegraphics[width=0.11\linewidth,trim={0cm 1.55cm 0 0},clip]{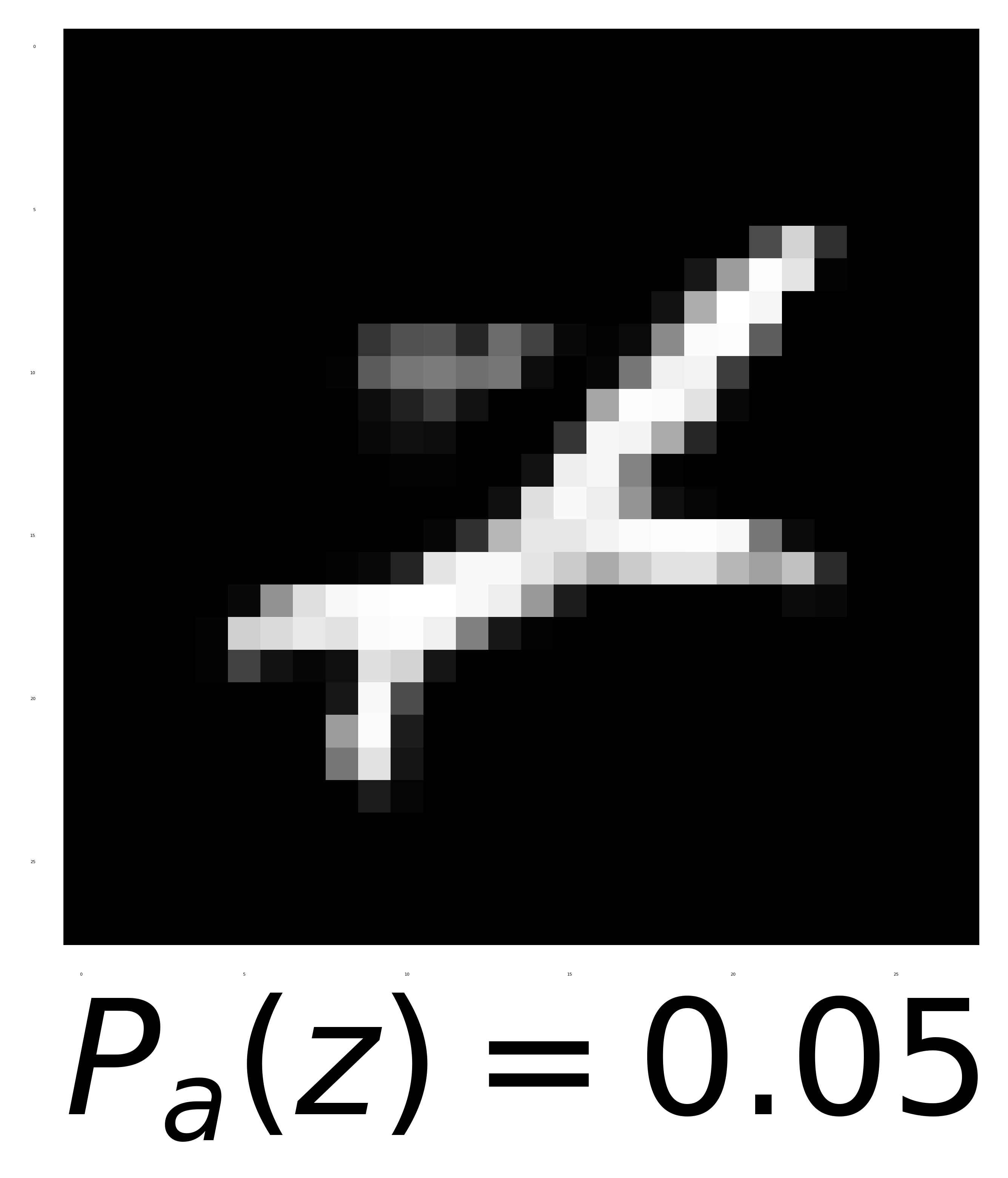}
    } 
    \subfloat[$P_a(z) = 0.68$]
    {   
        \includegraphics[width=0.11\linewidth,trim={0cm 1.55cm 0 0},clip]{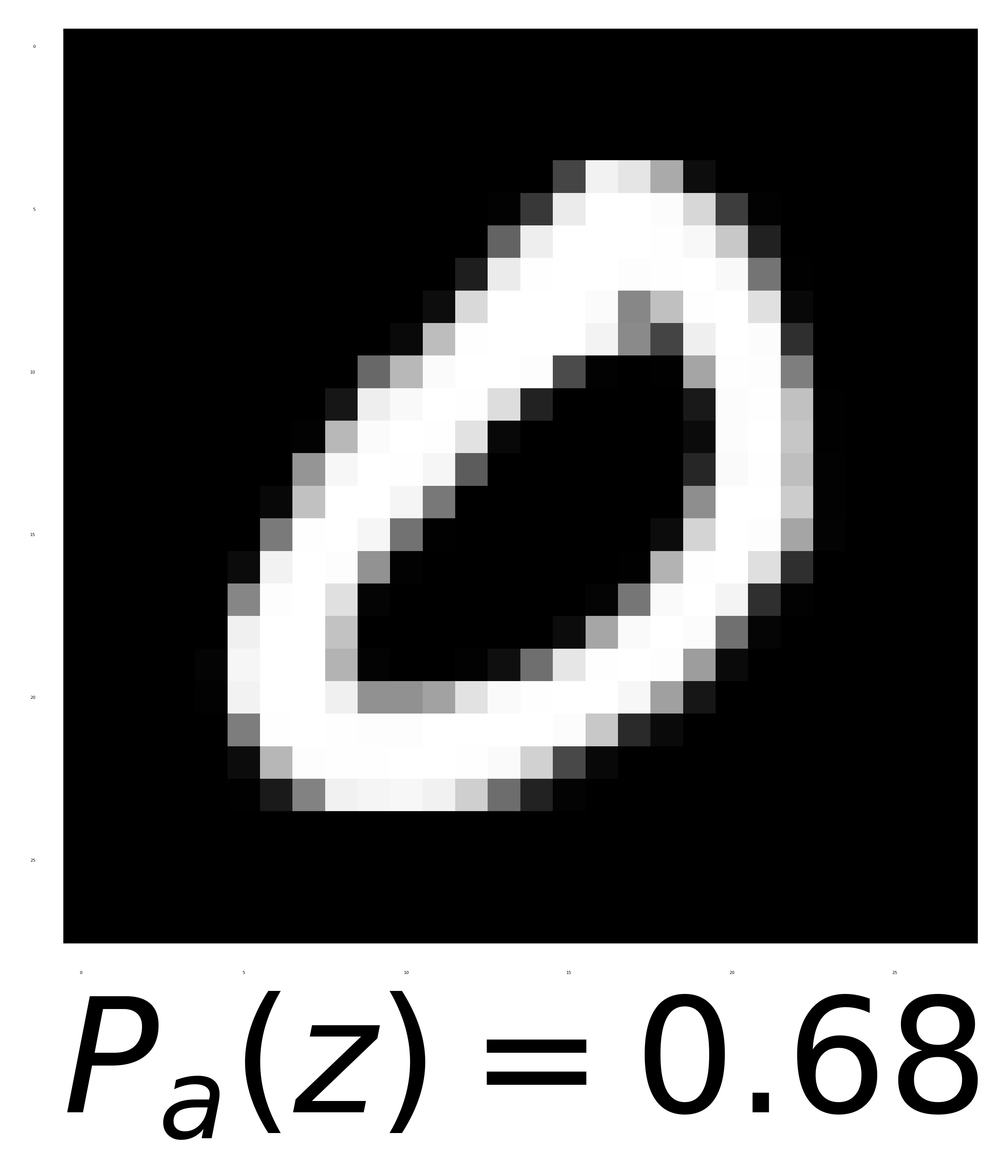}
    } 
    \subfloat[$P_a(z) = 0.73$]
    {   
        \includegraphics[width=0.11\linewidth,trim={0cm 1.55cm 0 0},clip]{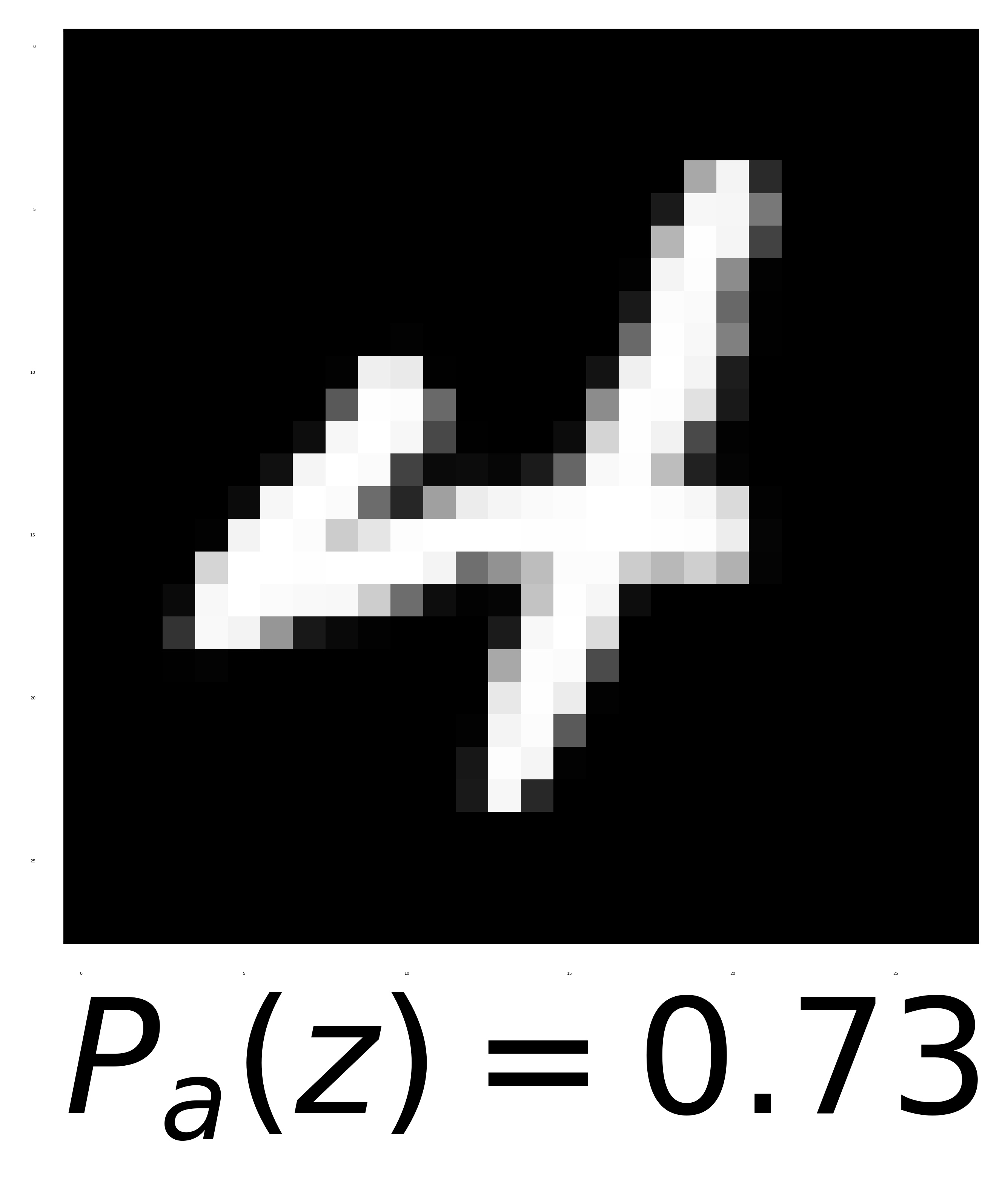}
    } 
    \subfloat[$P_a(z) = 0.00$]
    {   
        \includegraphics[width=0.11\linewidth,trim={0cm 1.55cm 0 0},clip]{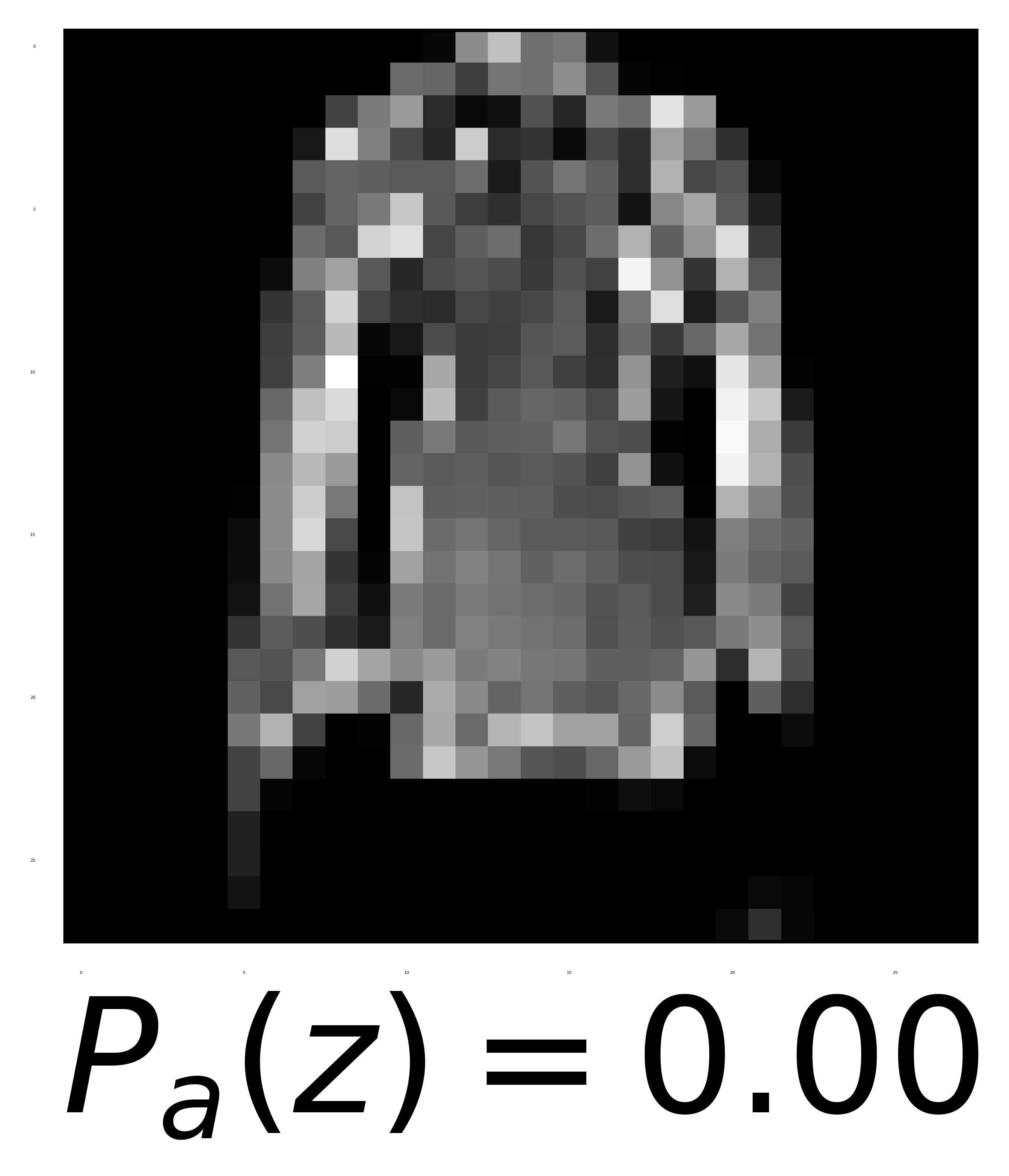}
    } 
    \subfloat[$P_a(z) = 0.08$]
    {   
        \includegraphics[width=0.11\linewidth,trim={0cm 1.55cm 0 0},clip]{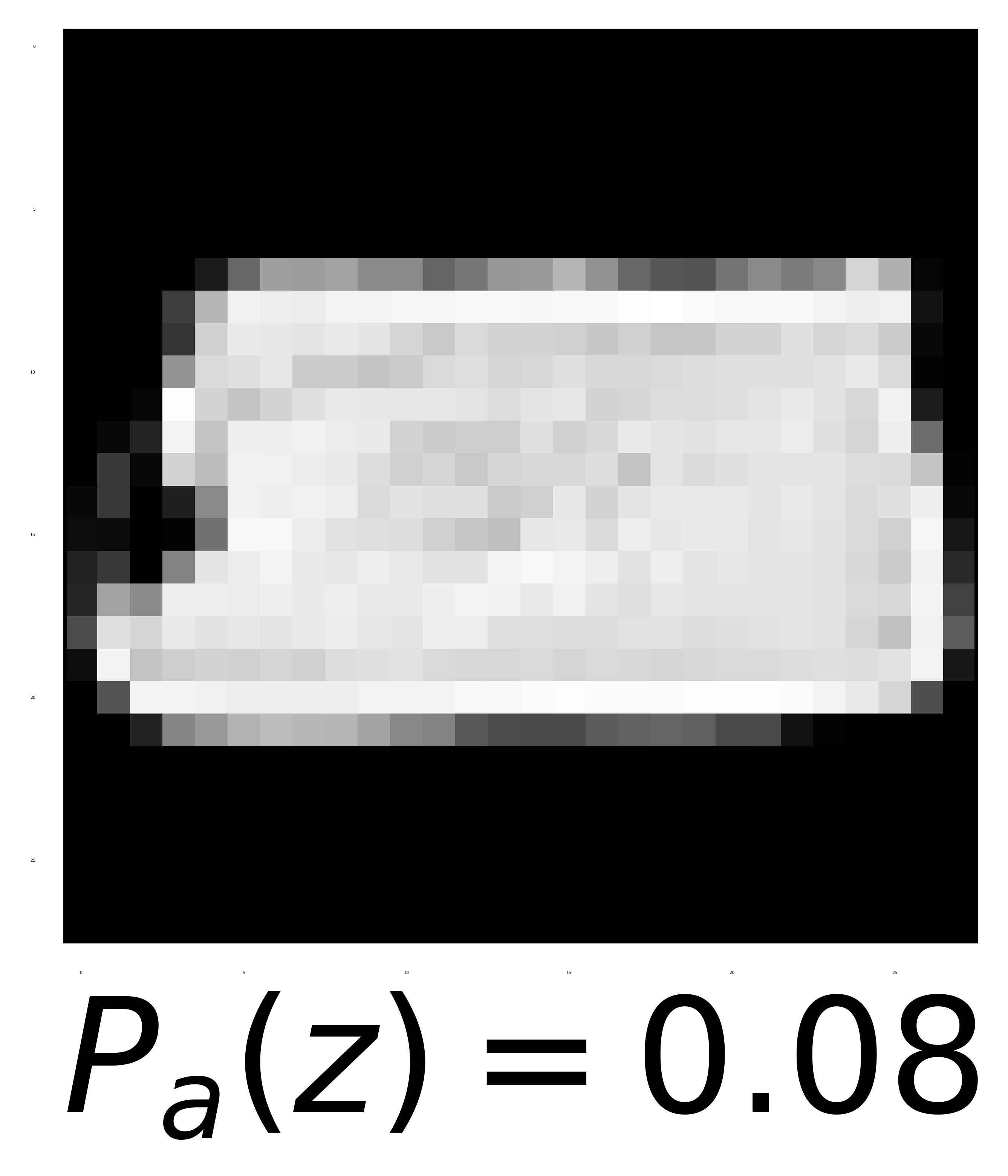}
    } 
    \subfloat[$P_a(z) = 0.57$]
    {   
        \includegraphics[width=0.11\linewidth,trim={0cm 1.55cm 0 0},clip]{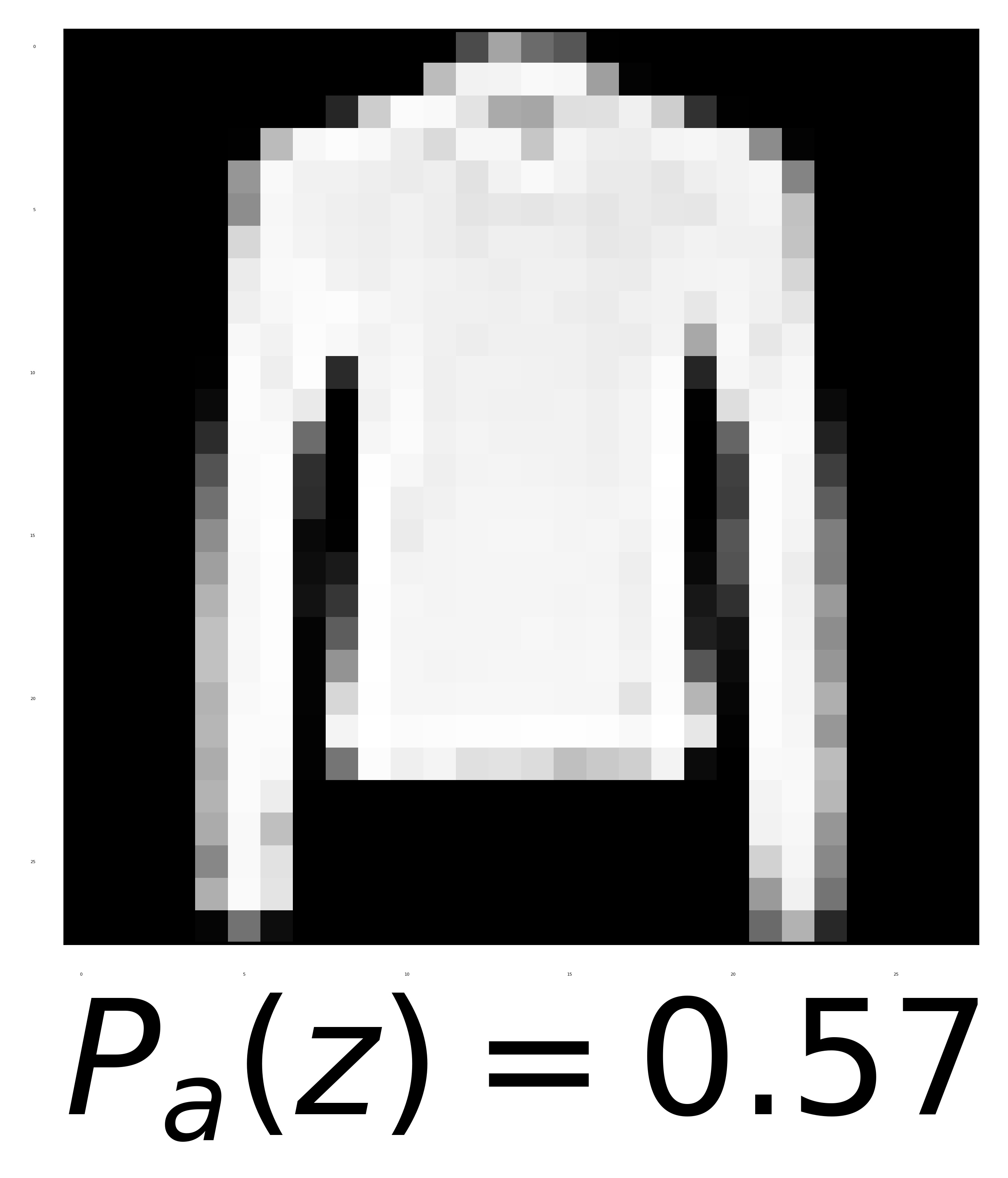}
    }
    \subfloat[$P_a(z) = 0.73$]
    {   
        \includegraphics[width=0.11\linewidth,trim={0cm 1.55cm 0 0},clip]{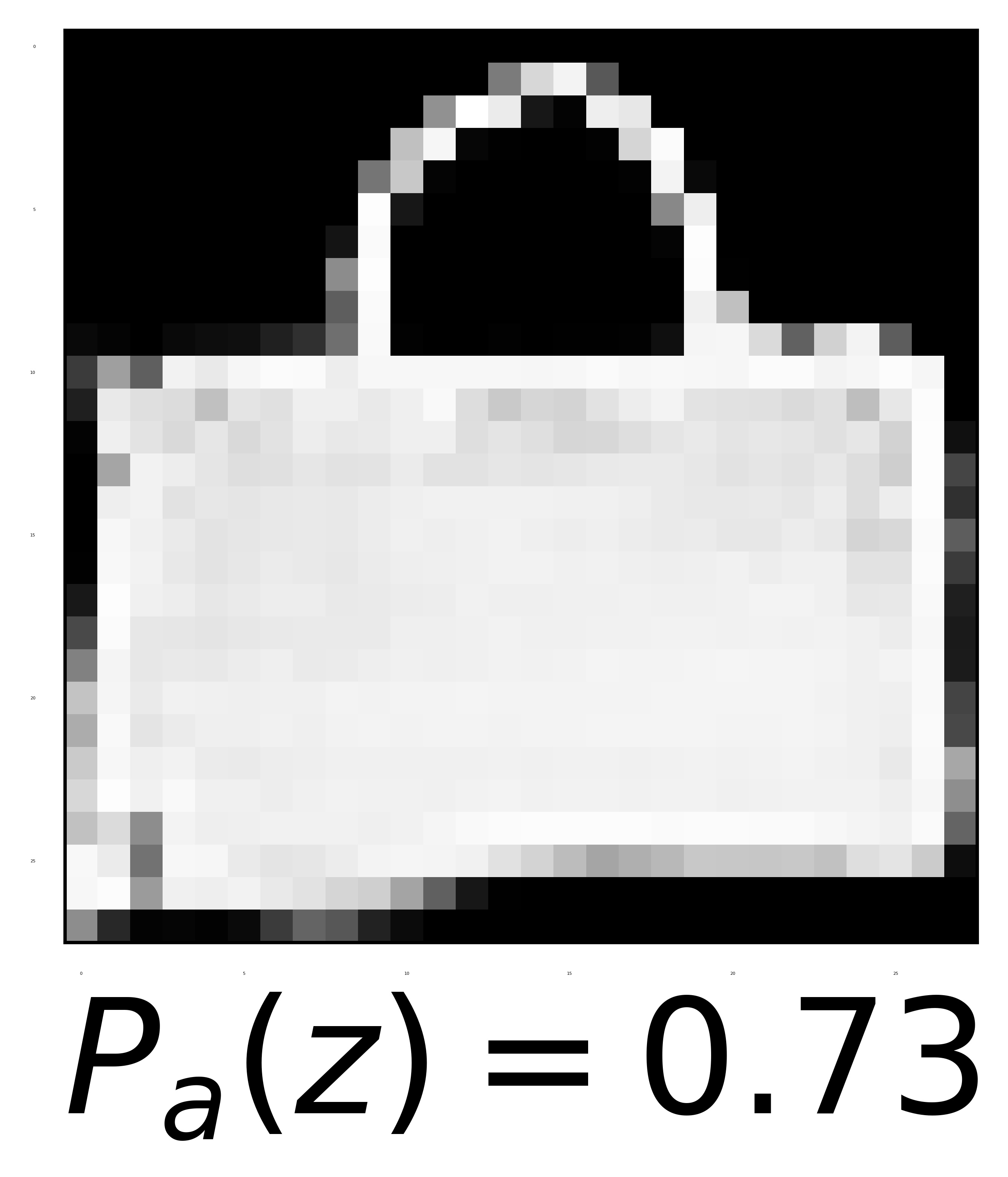}
    } \\
    \subfloat[$P_a(z) = 0.00$]
    {   
        \includegraphics[width=0.11\linewidth,trim={0cm 0.47cm 0 0},clip]{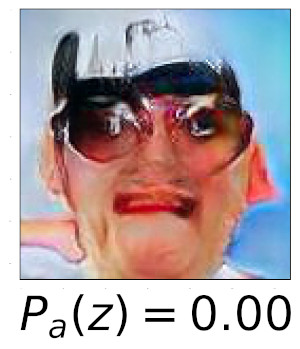}
    } 
    \subfloat[$P_a(z) = 0.04$]
    {   
        \includegraphics[width=0.11\linewidth,trim={0cm 0.47cm 0 0},clip]{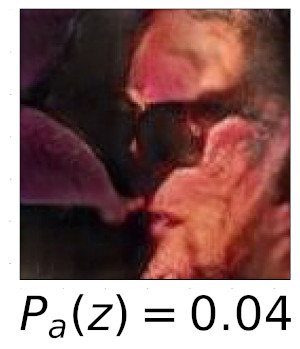}
    }
    \subfloat[$P_a(z) = 0.38$]
    {   
        \includegraphics[width=0.11\linewidth,trim={0cm 0.47cm 0 0},clip]{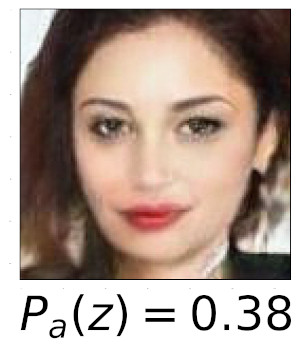}
    }
    \subfloat[$P_a(z) = 0.69$]
    {   
        \includegraphics[width=0.11\linewidth,trim={0cm 0.47cm 0 0},clip]{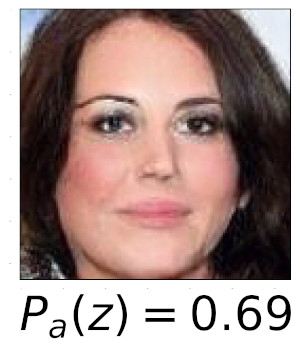}
    } 
    \subfloat[$P_a(z) = 0.00$]
    {   
        \includegraphics[width=0.11\linewidth,trim={0cm 0.18cm 0 0},clip]{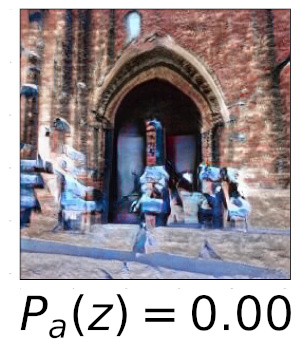}
    } 
    \subfloat[$P_a(z) = 0.06$]
    {   
        \includegraphics[width=0.11\linewidth,trim={0cm 0.18cm 0 0},clip]{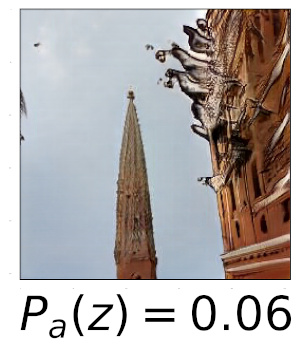}
    } 
    \subfloat[$P_a(z) = 0.58$]
    {   
        \includegraphics[width=0.11\linewidth,trim={0cm 0.18cm 0 0},clip]{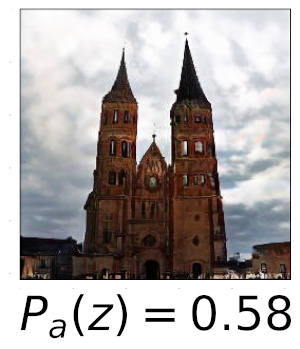}
    } 
    \subfloat[$P_a(z) = 0.69$]
    {   
        \includegraphics[width=0.11\linewidth,trim={0cm 0.18cm 0 0},clip]{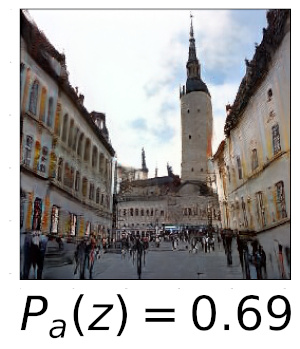}
    } 
    \\
    \caption{Images drawn from the generative model and their acceptance probabilities with the latentRS algorithm, given by the network $w^\varphi$. As expected, the quality of images correlates with higher acceptance rates on all datasets: MNIST, F-MNIST, CelebA, and LSUN.\label{fig:celeba_w5_t5}}
\end{figure*}
\captionsetup[subfigure]{labelformat=parens}

\begin{figure*}
    \centering
    \subfloat
    {   
        \includegraphics[width=0.9\linewidth]{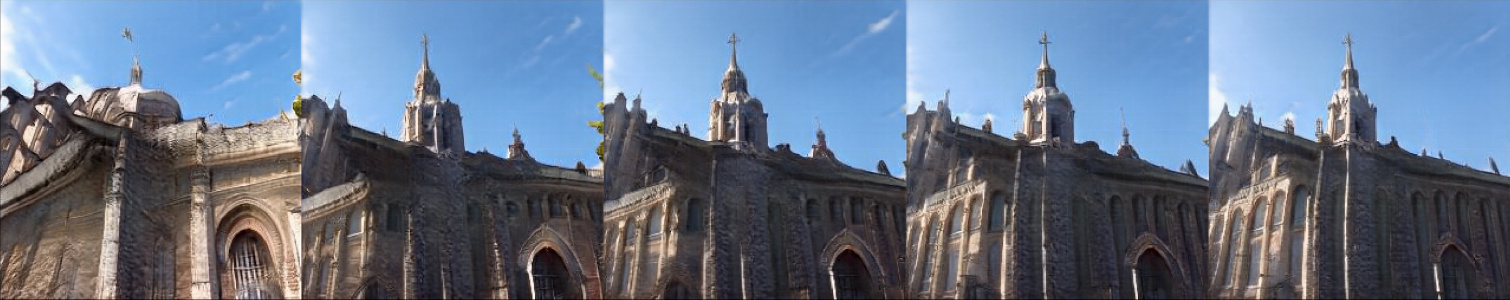}
    }
    \\
    \subfloat
    {   
        \includegraphics[width=0.9\linewidth]{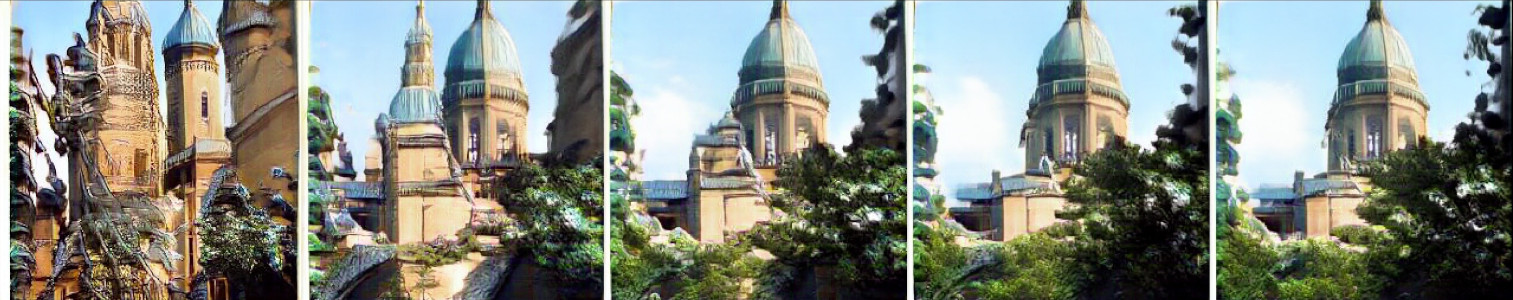}
    }
    \\
    \subfloat
    {   
        \includegraphics[width=0.9\linewidth]{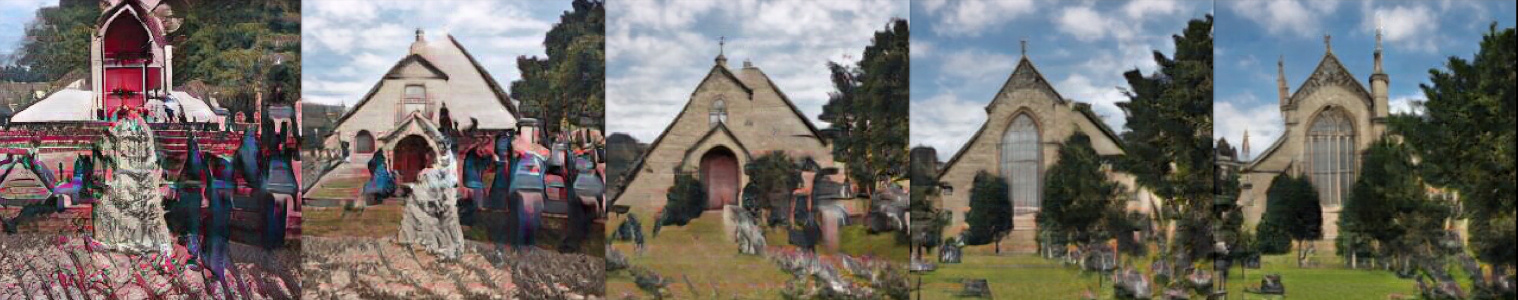}
    }
    \\
    \subfloat
    {   
        \includegraphics[width=0.3\linewidth]{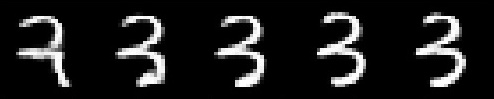}
    }\hspace{-0.2cm}
    \subfloat
    {   
        \includegraphics[width=0.3\linewidth]{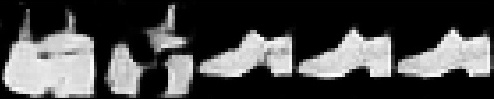}
    }\hspace{-0.2cm}
    \subfloat
    {   
        \includegraphics[width=0.3\linewidth]{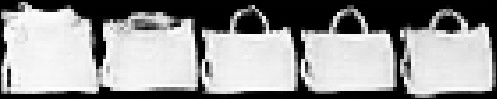}
    }\\
    \subfloat{\includegraphics[width=0.42\linewidth]{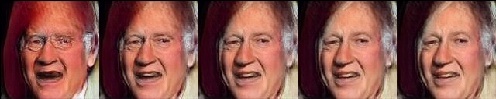}}\hspace{1.cm} 
    \subfloat{\includegraphics[width=0.42\linewidth]{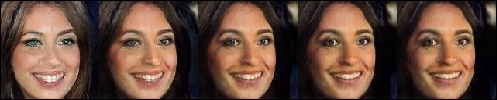}}
    \\
    \caption{Gradient ascent on latent importance weights (latentGA): the quality is gradually improved as we move to larger importance weights. Each image is generated only for visualization, and one can run this gradient ascent directly in the latent space using $w^\varphi$. Interestingly, this gradient ascent only involves a simple MLP network which is computationally cheap.  \label{fig:latent_gradient_descent} }
\end{figure*}

\subsubsection{Description of datasets and neural architectures} 
We first consider two well-known image datasets that are \textbf{MNIST} \cite{lecun98gradientbasedlearning} and  \textbf{FashionMNIST} (F-MNIST). We follow \cite{khayatkhoei2018disconnected} and use a standard CNN architecture composed of a sequence of blocks made of 3x3 convolution layer and ReLU activations with nearest neighbor upsampling. For these datasets, the discriminator is trained using the hinge loss \cite{lim2017geometric} with gradient penalty (Hinge-GP). Finally, the architecture used for the network $w^\varphi$ is very simple: an MLP with 4 fully-connected layers and ReLU activation (with a $\text{width} =4 \times d$).

\textbf{CelebA} \cite{liu2015faceattributes} is a large-scale dataset of faces covering a variety of poses. We use a pre-trained model of Progressive GAN \cite{karras2017progressive} at 128x128 resolution. The discriminator is trained using a Wasserstein loss with gradient-penalty. Also, the architecture used for the network $w^\varphi$ is really standard: a 5 hidden-layer MLP with a width of the same size than the latent space dimension.  

\textbf{LSUN Church} \cite{yu15lsun} is a dataset of church images with a lot of variety. We use a pre-trained model of StyleGAN2 \cite{karras2020analyzing} at 256x256 resolution. Similarly to the CelebA dataset, the discriminator is trained using a Wasserstein loss with gradient-penalty. Also, the architecture used for the network $w^\varphi$ is a 3 hidden-layer MLP with width equal to the latent space dimension. Note that the StyleGAN architecture already contains an 8-layer MLP network $M_\theta: \mathds{R}^d \to \mathds{R}^d$ that transforms a latent space variable to an intermediate latent variable \cite{karras2018style}. We consequently leverage this pre-trained $M_\theta$ and train the network $w^\varphi$ on top of it. 

\subsubsection{Results}
The main results of this comparison are shown in Table \ref{table:real_worlds_results_main} and Figure \ref{fig:comparison}. On all studied datasets, our latentRS+GA outperforms every other method on the EMD with lower computational cost. Interestingly, latentRS achieves good performance on FID while being more than 15 times faster. Figure \ref{fig:comparison} is particularly interesting since it gives a good visualization of the trade-off between computational cost and quality of the generated samples. On this experiment ran on CelebA and LSUN, we observe that latentRS+GA can achieve a significantly better precision than both SIR and DOT while being much faster. Interestingly, even though these datasets are high-dimensional, contain only one-class, and $w^\varphi$ has a low capacity, our proposed methods still produce interesting results. 

To visualize the efficiency of the proposed method, Figure \ref{fig:celeba_w5_t5} shows generated samples along with their acceptance probabilities. As expected, we observe that higher acceptance probabilities correlate with higher quality images. Figure \ref{fig:latent_gradient_descent} stresses how generated images improve when performing latent gradient ascent on the importance weights. Finally, we provide more qualitative results and details on the experiments in supplementary material. 

\section{Conclusion}
This paper deals with improving the quality of pre-trained GANs. Conversely, to concurrent methods which leverage the discriminator at inference time, we propose to train adversarially a neural network which learns importance weights in the latent space of GANs. These latent importance weights are then used with two complementary sampling methods: latentRS and latentGA. We experimentally show that this latent reweighting consistently enhances the quality of the pre-trained model. When these two methods are combined in latentRS+GA, it surpasses concurrent post-training methods while being less computationally intensive. 

\bibliography{main}
\clearpage
\appendix
\onecolumn

\section{Proof of Lemma \ref{lem:lemma1}}
Let's prove that $\mathds{E}_{\gamma^\varphi}=1$. We have that:
\begin{equation*}
    \int_{\mathds{R}^d} 1 \rm \gamma^\varphi({\rm d}z) = \int_{\mathds{R}^d} w^\varphi(z) \gamma({\rm d}z) =1,
\end{equation*}
by assumption. Consequently, the measure $\gamma^\varphi$ is a well-defined probability distribution on $\mathds{R}^d$.

\section{Proof of Theorem \ref{th:theorem1}}
It is clear that the network $w^\varphi$ is a density function with respect to the distribution $\gamma$ defined on $\mathds{R}^d$. Consequently, the measure $\mu_\theta^\varphi$ is absolutely continuous with respect to $\mu_\theta$ and thus with respect to the Lebesgue measure.

We start the proof by stating that for any absolutely continuous distribution $\mu_\theta^\varphi$, there exists an optimal transport $T_\theta^\varphi$, \cite[Theorem B]{pratelli2007equality}, such that $W(\mu_n, \mu_\theta^\varphi) = \int_{\mathds{R}^D} \|x-T_\theta^\varphi(x)\| \rm d\mu_\theta^\varphi$ \cite{hartmann2020semi}. Recall that for any $x \in \mathds{R}^D$, there exists $X_i \in [1,n]$ such that $T_\theta^\varphi(x)=X_i$. Since $\mu_\theta^\varphi$ is absolutely continuous, there exists a ball $B(z, r)$ centered in $z \in \mathds{R}^d$ with radius $r>0$ such that $\mu_\theta^\varphi(B(z, r)) > 0$ and, we have:
\begin{enumerate}
    \item there exists $i \in [1,n]$ such that for all $x \in B(z, r)$, $T_\theta^\varphi(x)=X_i$, 
    \item for all $x \in B(z, r)$, $\|x-X_i\| >\|X_i-\Tilde{X_i}\|$, recall that $\Tilde{X_i} = \underset{z \in \mathds{R}^d}{\argmin} \  \|X_i-G_\theta(z)\|$. 
\end{enumerate}
  
Consequently, we have: 
\begin{align*}
    W(\mu_n, \mu_\theta^\varphi) = \int_{\mathds{R}^D} \|x-T_\theta^\varphi(x)\| 
    \mu_\theta^\varphi({\rm d} x) &= \int_{\mathds{R}^D \backslash B(z, r)} \|x-T_\theta^\varphi(x)\| 
    \mu_\theta^\varphi({\rm d} x) + \int_{B(z,r)} \|x-T_\theta^\varphi(x)\| 
    \mu_\theta^\varphi({\rm d} x) \\
    &> \int_{\mathds{R}^D \backslash B(z, r)} \|x-T_\theta^\varphi(x)\| 
    \mu_\theta^\varphi({\rm d} x) + 
    \int_{B(z,r)} \|X_i-\Tilde{X_i}\| 
    \mu_\theta^\varphi({\rm d} x) \\
    & \geqslant \int_{\mathds{R}^D \backslash B(z, r)} \|\Tilde{T}_\theta^\varphi(x)-T_\theta^\varphi(x)\| \mu_\theta^\varphi({\rm d} x) + 
    \int_{B(z,r)} \|X_i-\Tilde{X_i}\| 
    \mu_\theta^\varphi({\rm d} x) \\
    & (\text{where $\Tilde{T}_\theta^\varphi(x) = \underset{z \in \mathds{R}^d}{\min} \ \|z-T_\theta^\varphi(x)\|$}) \\
    & = \frac{1}{n} \sum_{i=1}^n \|X_i - \Tilde{X_i} \| \\
    &= W(\mu_n, \frac{1}{n} \sum_{i=1}^n \delta(\Tilde{X_i})),
\end{align*}
where $\delta$ refers to the Dirac probability distribution.

Finally, when taking the infinimum over all continuous functions $\varphi$, we have that:
\begin{equation*}
    W(\mu_n, \frac{1}{n} \sum_{i=1}^n \delta(\Tilde{X_i})) \leqslant \underset{\varphi \in \Phi}{\inf} W(\mu_n, \mu_\theta^\varphi)
\end{equation*}

\section{Evaluation details}
\textbf{Precision recall metric. } For the precision-recall metric, we use the algorithm from \cite{khayatkhoei2018disconnected}. Namely, when comparing the set of real data points $(x_1,...,x_n)$ with the set of fake data points $(y_1,...,y_n)$:

A point $x_i$ has a recall $r(x_i) = 1$ if there exists $y_j$, such that $\| x_i - y_j \| \leq \| y_j - y_j(k) \| $, where $y_j(k)$ is the k-nearest neighbor of n. Finally, the recall is the average of individual recall: $\frac{1}{n} \sum_i r(x_i) $.

A point $y_i$ has a precision $p(y_i) = 1$ if there exists $x_j$, such that $\| y_i - x_j \| \leq \| x_j - x_j(k) \| $, where $x_j(k)$ is the k-nearest neighbor of n. Finally, the precision is the average of individual precision: $\frac{1}{n} \sum_i p(x_i) $.

\textbf{Parameters. } For all datasets, we use $k=3$ (3rd nearest neighbor). For MNIST and F-MNIST, we use a set of $n=2048$ points. For CelebA and LSUN Church, we use a set of $n=1024$ points. This is also valid for the EMD. For FID, we use the standard protocol with $n=50000$ points and Inception Net. We run 10 evaluations of each metric (each evaluation is done with a different set of random points), report the average and the 97\% confidence interval by considering that we have 10 i.i.d. samples from a normal distribution. 

\section{Sampling algorithms: latentRS, latentGA, and latentRS+GA}
We present here the three sampling algorithms associated with our importance weight function $w^\varphi$. See Section \ref{section:hyper-params} for details on the hyper-parameters used in latentGA and latentRS+GA. 
\begin{algorithm}
\SetAlgoLined
 \textbf{Requires:} Prior $Z$, Gen. $G_\theta$, Importance weight network $w^\varphi$, maximum importance weight $m$\;
 \While{True}{
  Sample $z \sim Z$ \;
  Sample $\alpha \sim \text{Uniform}[0,1]$ \;
  \If{$\frac{w^\varphi(z)}{m} \geq \alpha$}{
   break\;
   }
 }
 $x \gets G_\theta(z)$\;
 \KwResult{Selected point x}
 \caption{LatentRS \label{algo:lrs_app}}
\end{algorithm}

\begin{algorithm}
\SetAlgoLined
 \textbf{Requires:} Prior $Z$, number of dimensions of the prior $d$, Gen. $G_\theta$, Importance weight network $w^\varphi$, number of steps $N$, step size $\varepsilon$\;
 Sample $z \sim Z$ \;
 \For{$n=1:N$}{
  $\text{grad}_z \gets \nabla_z w^\varphi(z)$ \;
  \If{$Z == \mathcal{N}(0,I)$ \textit{\textbf{and}} $d>>1$}{
  \#\# Projection step for high-dimensional gaussians  \#\# \;
  $\text{grad}_z \gets \text{grad}_z - (\text{grad}_z \cdot z) z / \sqrt{d} $ \; \hspace{0.3cm} 
  }
  $z  \gets z + \varepsilon * \text{grad}_z$ \;
  
 }
 $x \gets G_\theta(z)$\;
 \KwResult{Selected point x}
 \caption{Latent Gradient Ascent (latentGA)}
\end{algorithm}

\begin{algorithm}
\SetAlgoLined
 \textbf{Requires:} Prior  $Z$, Number of dimensions of the prior $d$, Gen. $G_\theta$, Importance weight network $w^\varphi$, maximum importance weight $m$, number of steps $N$, step size $\varepsilon$\;
 \While{True}{
  Sample $z \sim Z$ \;
  Sample $\alpha \sim \text{Uniform}[0;1]$ \;
  \If{$\frac{w^\varphi(z)}{m} \geq \alpha$}{
   break\;
   }
 }
 \For{$n=1:N$}{
  $\text{grad}_z \gets \nabla_z w^\varphi(z)$ \;
  \If{$Z = \mathcal{N}(0,I)$ \textit{\textbf{and}} $d>>1$}{
  \#\# Projection step for high-dimensional gaussians \#\# \;
  $\text{grad}_z \gets \text{grad}_z - (\text{grad}_z \cdot z) z / \sqrt{d} $ \; \hspace{0.3cm} 
  }
  $z  \gets z + \varepsilon * \text{grad}_z$ \;
  
 }
 Compute $x = G_\theta(z)$\;
 \KwResult{Selected point x}
 \caption{Latent RS+GA}
\end{algorithm}

\section{Hyper-parameters. \label{section:hyper-params}}

\textbf{SIR \cite{grover2019bias}:} Model selection: we fine-tune with a binary cross-entropy loss the discriminator from the end of the adversarial training and select the best model in terms of EMD. We tested with/without regularizing the discriminator during the fine-tuning (with gradient penalty or spectral normalization). Without regularization, the performance drops fast. Best results are obtained by regularizing the discriminator, thus we report these results.

We use then use Sampling-Importance-Resampling algorithm. In SIR, we sample $N$ points from the generator, compute their importance weights according density ratios, and accept one of them (each point is accepted with a probability proportional to its importance weight). The hyper-parameter of SIR algorithm is $N$. Results for grid search on $N$ are shown below in Table \ref{ablation_progan} and Table \ref{ablation_style_gan2}. In Table \ref{table:real_worlds_results}, results are shown with $N = 10$.

\textbf{DOT \cite{tanaka2019discriminator}:} Model selection: we fine-tune with the WGAN-GP loss the discriminator from the end of the adversarial training and select the best model in terms of EMD, when running DOT. We perform a projected gradient descent as described in \cite{tanaka2019discriminator} with SGD.  Hyper-parameters are the number of steps $N_{steps}$ and the step size $\varepsilon$. We made the following grid search: $N_{steps} = [2,5,10,50]$ and $\varepsilon = [0.01,0.05,0.1]$. Results for grid search on $N_{steps}$ are shown below in Table \ref{ablation_progan} and Table \ref{ablation_style_gan2}. In Table \ref{table:real_worlds_results}, results are shown with $N_{steps} = 10$ and $\varepsilon = 0.05$ or $\varepsilon = 0.01$ depending on the dataset (we select the best one).

\textbf{Training of $w^\varphi$:}
For MNIST and F-MNIST, we use the same hyper-parameters: $\lambda_1 = 10$, $\lambda_2 = 3$ and $m = 3$. $w^\varphi$ is a standard MLP with 4 hidden layers, each having 400 nodes (4x dimension of latent space), and relu activation. The output layer is 1-dimensional and with a relu activation. The learning rate of the discriminator is $4*10^{-4}$, the learning rate of $w^\varphi$ is $10^{-4}$. The two networks are optimized with Adam algorithm, where we set $\beta = (0.5,0.5)$. We use 1 step of importance weight optimization for 1 step of discriminator optimization.  

For Progressive GAN on CelebA (128x128), we use: $\lambda_1 = 20$, $\lambda_2 = 5$ and $m = 3$. $w^\varphi$ is a standard MLP with 4 hidden layers, each having 512 nodes (1x dimension of latent space), and leaky-relu activation (0.2 of negative slope). The output layer is 1-dimensional and with relu activation. Since we do not have the pre-trained discriminator, we first train a WGAN-GP discriminator between ProGAN and CelebA images for 500 steps, and then start the adversarial training of $w^\varphi$. The learning rate of the discriminator is $10^{-4}$, the learning rate of $w^\varphi$ is $10^{-5}$. The two networks are optimized with Adam algorithm, where we set $\beta = (0.,0.999)$. During optimization, we perform iteratively 3 $w^\varphi$ updates and 1 discriminator's updates.

For StyleGAN2 on LSUN Church (256x256), we use: $\lambda_1 = 30$, $\lambda_2 = 5$ and $m = 2$. $w^\varphi$ is a standard MLP with 3 hidden layers, each having 512 nodes (1x dimension of latent space), and leaky-relu activation (0.2 of negative slope). The output layer is 1-dimensional and with a relu activation. Since we do not have the pre-trained discriminator, we first train a WGAN-GP discriminator between StyleGAN2 and LSUN Church images for 500 steps, and then start the adversarial training of $w^\varphi$. The learning rate of the discriminator is $10^{-4}$, the learning rate of $w^\varphi$ is $10^{-5}$. The two networks are optimized with Adam algorithm, where we set $\beta = (0.,0.999)$. During optimization, we perform 3 $w^\varphi$ updates for 1 discriminator's updates. 

\textbf{LatentRS:} Once the network $w^\varphi$ is trained (see above), there is no hyper-parameter for latentRS algorithm.

\textbf{LatentGA and latentRS+GA:} We use the same neural network than in LatentRS. The hyper-parameters for this method are similar to DOT: number of steps of gradient ascent $N_{steps}$ and step size $\varepsilon$. With the model selected on LRS, we make the following grid search: $N_{steps} = [2,5,10,50]$ and $\varepsilon = [0.01,0.05,0.1]$. Best results were obtained with $\varepsilon=0.05$ on all datasets. Results for grid search on $N_{steps}$ are shown below in Table \ref{ablation_progan} and Table \ref{ablation_style_gan2}. In Table \ref{table:real_worlds_results_main}, results are shown with $N_{steps} = 10$ and $\varepsilon = 0.05$.

\clearpage 
\newpage

\section{Comparisons with concurrent methods on synthetic and real-world datasets}
In this section, we provide more quantitative results: a comparison of SIR, DOT, SIR, LatentRS and latentRS+GA on MNIST and F-MNIST in Table \ref{table:real_worlds_results_MNIST}; an ablation study on the impact of number of points (respectively gradient ascent steps) in SIR (respectively DOT, latentGA and latentRS+GA), on ProGAN trained on CelebA in Table \ref{ablation_progan} and StyleGAN2 trained on Lsun Church in Table \ref{ablation_style_gan2}. \\

\begin{table}[h]
\begin{center}
\begin{tabular}{|l|c|c|c|c|c|}
\cline{2-6}
\multicolumn{1}{l|}{\textbf{MNIST}}  & Prec. ($\uparrow$) & Rec. ($\uparrow$) &  EMD ($\downarrow$) & FID ($\downarrow$) & Inference (ms) \\
\hline
Hinge-GP & $87.4 {\scriptstyle \pm 0.9}$ & $94.6 {\scriptstyle \pm 0.4} $ & $24.9 {\scriptstyle  \pm 0.3}$ & $53.6 {\scriptstyle  \pm 7.2}$ & $0.7$\\ \hline
HGP: SIR & $88.8{\scriptstyle \pm 1.0}$ & $94.3 {\scriptstyle \pm 0.5}$ & $24.2 {\scriptstyle \pm 0.2}$ &  ${38.7 {\scriptstyle \pm 3.1}}$& $10.0$\\ 
HGP: DOT & $89.5 {\scriptstyle \pm 0.6}$ & $94.0 {\scriptstyle  \pm 0.3}$ & $ 24.8 {\scriptstyle  \pm 0.2}$ & $43.3 {\scriptstyle  \pm 3.4}$  & $15.7$\\ 
HGP: latentRS ($\star$) & $89.0 {\scriptstyle \pm 0.4}$ & $\mathbf{94.7 {\scriptstyle  \pm 0.7}}$ & $24.1 {\scriptstyle  \pm 0.3}$ & $\mathbf{36.3 {\scriptstyle  \pm 3.2}}$ & $\mathbf{1.6}$\\ 
HGP: latentRS+GA ($\star$) & $\mathbf{91.8 {\scriptstyle \pm 1.0}}$ & $92.8 {\scriptstyle  \pm 0.4}$ & $\mathbf{23.4 {\scriptstyle  \pm 0.2}}$ & $38.2 {\scriptstyle  \pm 3.8}$ & $8.6$\\ 
\hline
\multicolumn{4}{l} {\textbf{F-MNIST}}  \\
\hline
Hinge-GP & $86.4 {\scriptstyle \pm 0.6}$ &  $86.8 {\scriptstyle \pm 0.6}$ & $68.6 {\scriptstyle \pm 0.4}$ & $598.9 {\scriptstyle \pm 55.5}$ & $0.7$\\ \hline 
HGP: SIR & $86.6 {\scriptstyle \pm 1.1}$ & $\mathbf{88.0 {\scriptstyle \pm 0.5}}$ & $68.0 {\scriptstyle \pm 0.5}$ &  $499.6 {\scriptstyle \pm 31.1}$ & $10.0$\\ 
HGP: DOT & $\mathbf{88.7 {\scriptstyle \pm 0.6}}$ & $86.6 {\scriptstyle  \pm 0.7}$ & $67.7 {\scriptstyle  \pm 0.5}$ & $508.3 {\scriptstyle  \pm 45.7}$ &  $15.7$\\ 
HGP: latentRS ($\star$) & $86.8 {\scriptstyle \pm 0.8}$ & ${87.5 {\scriptstyle \pm 0.9}}$ & ${67.6 {\scriptstyle \pm 0.6}}$ & $\mathbf{438.3 {\scriptstyle \pm 50.2}}$ & $\mathbf{1.6}$\\ 
HGP: latentRS+GA ($\star$) & ${88.4 {\scriptstyle \pm 0.7}}$ & ${86.8 {\scriptstyle \pm 0.7}}$ & $\mathbf{67.0 {\scriptstyle \pm 0.9}}$ & ${475.5 {\scriptstyle \pm 58.5}}$ & $8.6$\\
\hline
\end{tabular}
\end{center}
\caption{latentRS+GA is the best performer and latentRS matches SOTA with a significantly reduced inference cost (by an order of at least 10). FID was computed using the same dataset-specific classifier used for the Precision/Recall metric. $\pm$ is $97\%$ confidence interval. Inference refers to the time in milliseconds needed to compute one image on a NVIDIA V100 GPU. \label{table:real_worlds_results_MNIST}}
\end{table}

\begin{table*}[h]
\begin{center}
\begin{tabular}{|l|c|c|c|c|}
\cline{2-5}
\multicolumn{1}{l|}{\textbf{CelebA 128x128}} & Precision & Recall & EMD  & Inference Time \\
\hline
ProGAN & $74.2 {\scriptstyle \pm 0.9}$ &  $60.7 {\scriptstyle \pm 1.4}$ & $25.4 {\scriptstyle \pm 0.1}$  & $3.6$\\ \hline
ProGAN: SIR (n=2) & ${78.2 {\scriptstyle \pm 1.0}}$ &  $58.4 {\scriptstyle \pm 1.3}$ & $25.0 {\scriptstyle \pm 0.1}$  & $9.8$\\ 
ProGAN: SIR (n=5) & ${79.3 {\scriptstyle \pm 0.6}}$ &  $57.6 {\scriptstyle \pm 1.3}$ & $24.9 {\scriptstyle \pm 0.1}$ & $24.5$\\ 
ProGAN: SIR (n=10) & ${79.5 {\scriptstyle \pm 0.4}}$ &  $57.3 {\scriptstyle \pm 1.0}$ & $24.9 {\scriptstyle \pm 0.2}$ & $49.0$\\ 
ProGAN: SIR (n=50) & ${80.2 {\scriptstyle \pm 1.0}}$ &  $57.4 {\scriptstyle \pm 1.4}$ & $25.0 {\scriptstyle \pm 0.1}$ & $245.0$\\ \hline
ProGAN: DOT (n=2) & ${78.2 {\scriptstyle \pm 1.1}}$ &  $58.6 {\scriptstyle \pm 1.1}$ & $24.9 {\scriptstyle \pm 0.1}$ &  $16.4$\\
ProGAN: DOT (n=5) & ${80.0 {\scriptstyle \pm 1.0}}$ &  $56.0 {\scriptstyle \pm 1.1}$ & $24.8 {\scriptstyle \pm 0.1}$ & $35.6$ \\
ProGAN: DOT (n=10) & ${81.3 {\scriptstyle \pm 1.0}}$ &  $52.9 {\scriptstyle \pm 1.4}$ & $25.0 {\scriptstyle \pm 0.1}$ & $67.6$\\
ProGAN: DOT (n=50) & ${82.3 {\scriptstyle \pm 0.7}}$ &  $52.1 {\scriptstyle \pm 1.3}$ & $25.0 {\scriptstyle \pm 0.1}$ &  $323.6$\\ 
\hline
ProGAN: latentGA (n=2) ($\star$) & ${76.7 {\scriptstyle \pm 1.2}}$ &  $\mathbf{59.4 {\scriptstyle \pm 0.9}}$ & $25.2 {\scriptstyle \pm 0.1}$ &$5.2$\\ 
ProGAN: latentGA (n=5) ($\star$) & ${77.8 {\scriptstyle \pm 1.2}}$ &  $58.4 {\scriptstyle \pm 0.7}$ & $25.1 {\scriptstyle \pm 0.1}$ & $7.6$\\ 
ProGAN: latentGA (n=10) ($\star$) & ${78.9 {\scriptstyle \pm 1.2}}$ &  $57.4 {\scriptstyle \pm 0.7}$ & ${25.0 {\scriptstyle \pm 0.1}}$ &  $11.6$ \\ 
ProGAN: latentGA (n=50) ($\star$) & ${84.1 {\scriptstyle \pm 1.2}}$ &  $49.0 {\scriptstyle \pm 1.3}$ & $24.8 {\scriptstyle \pm 0.1}$  & $43.6$\\ \hline
ProGAN: latentRS+GA (n=2) ($\star$) & ${81.2 {\scriptstyle \pm 0.8}}$ &  $55.3 {\scriptstyle \pm 1.5}$ & $24.7 {\scriptstyle \pm 0.1}$   &$6.1$\\ 
ProGAN: latentRS+GA (n=5) ($\star$) & ${82.1 {\scriptstyle \pm 0.7}}$ &  $54.3 {\scriptstyle \pm 1.2}$ & $24.6 {\scriptstyle \pm 0.2}$  & $8.5$\\ 
ProGAN: latentRS+GA (n=10) ($\star$) & ${83.3 {\scriptstyle \pm 1.0}}$ &  $52.7 {\scriptstyle \pm 0.9}$ & $\mathbf{24.5 {\scriptstyle \pm 0.1}}$ & $12.5$ \\ 
ProGAN: latentRS+GA (n=50) ($\star$) & $\mathbf{89.2 {\scriptstyle \pm 0.8}}$ &  $36.1 {\scriptstyle \pm 0.7}$ & $25.0 {\scriptstyle \pm 0.1}$& $44.5$\\ \hline
ProGAN: latentRS ($\star$) & $79.3 { {\scriptstyle \pm 1.0}}$ &  $56.5 {\scriptstyle \pm 1.2}$ & $24.8 {\scriptstyle \pm 0.2}$ & $\mathbf{4.5}$\\ 
\hline
\end{tabular}
\end{center}
\caption{Comparison of the proposed methods (latentRS, latentGA, and latentRS+GA) with concurrent methods on ProgressiveGan (CelebA 128x128). For this specific study, we explore different computational budgets for SIR, DOT, latentGA, and latentRS+GA. latentRS+GA enables a consistent gain in both EMD and precision for a reasonable computational overhead. \label{ablation_progan}}
\end{table*}

\begin{table*}[h]
\begin{center}
\begin{tabular}{|l|c|c|c|c|}
\cline{2-5}
\multicolumn{1}{l|}{\textbf{LSUN Church (256x256)}} & Precision & Recall & EMD & Inference Time \\
\hline
StyleGAN2 & $55.6 {\scriptstyle \pm 1.2}$ &  ${62.4 {\scriptstyle \pm 1.1}}$ & ${23.6 {\scriptstyle \pm 0.1}}$ & $11.7$ \\ \hline
StyleGAN2: SIR (n=2) & ${58.5 {\scriptstyle \pm 0.7}}$ &  $60.7 {\scriptstyle \pm 1.3}$ & $23.5 {\scriptstyle \pm 0.1}$ & $26.0$ \\ 
StyleGAN2: SIR (n=5) & ${59.8 {\scriptstyle \pm 1.1}}$ &  $59.0 {\scriptstyle \pm 1.2}$ & $23.5 {\scriptstyle \pm 0.1}$ & $65.0$ \\ 
StyleGAN2: SIR (n=10) & $60.5 {\scriptstyle \pm 1.4}$ & $58.1 {\scriptstyle \pm 1.3}$ & $23.4 {\scriptstyle \pm 0.1}$ & $130.0$\\
StyleGAN2: SIR (n=50) & ${61.2 {\scriptstyle \pm 1.2}}$ &  $57.8 {\scriptstyle \pm 0.9}$ & $23.4 {\scriptstyle \pm 0.1}$ & $650.0$ \\ \hline
StyleGAN2: DOT (n=2) & ${60.4 {\scriptstyle \pm 1.4}}$ &  $57.0 {\scriptstyle \pm 1.1}$ & $23.4 {\scriptstyle \pm 0.1}$ & $48.7$ \\
StyleGAN2: DOT (n=5) & ${64.1 {\scriptstyle \pm 0.9}}$ &  $52.2 {\scriptstyle \pm 1.0}$ & $23.2 {\scriptstyle \pm 0.1}$ & $104.2$ \\
StyleGAN2: DOT (n=10) & $67.4 {\scriptstyle \pm 1.4}$ & $48.3 {\scriptstyle \pm 1.0}$ & $23.1 {\scriptstyle \pm 0.1}$ & $196.7$ \\ 
StyleGAN2: DOT (n=50) & ${68.8 {\scriptstyle \pm 0.9}}$ &  $37.0 {\scriptstyle \pm 1.1}$ & $23.6 {\scriptstyle \pm 0.1}$ & $937.7$ \\ \hline
StyleGAN2: latentGA (n=2) ($\star$) & ${58.2 {\scriptstyle \pm 1.0}}$ &  $\mathbf{61.4 {\scriptstyle \pm 1.2}}$ & $23.4 {\scriptstyle \pm 0.1}$ & $17.1$ \\
StyleGAN2: latentGA (n=5) ($\star$) & ${61.1 {\scriptstyle \pm 0.9}}$ &  $58.5 {\scriptstyle \pm 1.1}$ & $23.2 {\scriptstyle \pm 0.1}$ & $25.2$ \\ 
StyleGAN2: latentGA (n=10) ($\star$) & ${64.6 {\scriptstyle \pm 0.9}}$ &  $55.9 {\scriptstyle \pm 1.5}$ & ${23.0 {\scriptstyle \pm 0.1}}$ & $38.7$ \\ 
StyleGAN2: latentGA (n=50) ($\star$) & ${69.9 {\scriptstyle \pm 1.1}}$ &  $47.2 {\scriptstyle \pm 1.4}$ & $22.8 {\scriptstyle \pm 0.1}$ & $146.7$\\ \hline
StyleGAN2: latentRS+GA (n=2) ($\star$) & ${66.3 {\scriptstyle \pm 1.2}}$ &  $54.8 {\scriptstyle \pm 1.3}$ & $23.0 {\scriptstyle \pm 0.1}$ & $21.6$\\ 
StyleGAN2: latentRS+GA (n=5) ($\star$) & ${69.6 {\scriptstyle \pm 1.0}}$ &  $50.6 {\scriptstyle \pm 0.9}$ & $22.8 {\scriptstyle \pm 0.2}$ & $29.7$ \\ 
StyleGAN2: latentRS+GA (n=10) ($\star$) & ${72.6 {\scriptstyle \pm 1.1}}$ &  $43.2 {\scriptstyle \pm 1.3}$ & $\mathbf{22.6 {\scriptstyle \pm 0.1}}$ & $43.2$\\ 
StyleGAN2: latentRS+GA (n=50) ($\star$) & $\mathbf{78.6 {\scriptstyle \pm 1.2}}$ &  $34.1 {\scriptstyle \pm 0.9}$ & $\mathbf{22.6 {\scriptstyle \pm 0.1}}$ & $151.2$ \\ \hline
StyleGAN2: latentRS ($\star$) & $63.3{ {\scriptstyle \pm 0.7}}$ &  $57.7 {\scriptstyle \pm 1.0}$ & $23.1 {\scriptstyle \pm 0.2}$ & $\mathbf{16.2}$\\ 
\hline
\end{tabular}
\end{center}
\caption{Comparison of the proposed methods (latentRS, latentGA, and latentRS+GA) with concurrent methods on StyleGAN2 (LSUN Church 256x256). For this specific study, we explore different computational budgets for SIR, DOT, latentGA, and latentRS+GA. latentRS+GA enables a consistent gain in both EMD and precision with a reasonable computational overhead. \label{ablation_style_gan2}}
\end{table*}

\begin{table}[h]
\begin{center}
\begin{tabular}{|l|c|}
\hline
\multicolumn{1}{|l|}{\textbf{MNIST/F-MNIST 28x28}} &  Inference Time \\
\hline
MNIST Generator & $0.7$ \\
MNIST Generator + Discriminator & $1.0$ \\ 
MNIST $\nabla_z D(G(z))$ (gradient for latent DOT) & $1.5$ \\ 
MNIST Network $w^\varphi$ ($\star$) & $0.3$ \\ 
MNIST: $\nabla_z W(z)$ (gradient for latent GA on IW) ($\star$) & $0.7$ \\ 
\hline
\hline
\multicolumn{1}{|l|}{\textbf{CelebA 128x128}} &  Inference Time \\
\hline
ProGAN Generator & $3.6 $ \\
ProGAN Generator + Discriminator & ${4.9 }$ \\ 
ProGAN $\nabla_z D(G(z))$ (gradient for latent DOT) & ${6.4 }$ \\ 
ProGAN Network $w^\varphi$ ($\star$) & ${0.3 }$ \\ 
ProGAN: $\nabla_z W(z)$ (gradient for latent GA on IW) ($\star$) & $0.8{ }$ \\ 
\hline
\hline
\multicolumn{1}{|l|}{\textbf{LSUN Church 256x256}} &  Inference Time \\
\hline
StyleGAN2 Generator & $11.7 $ \\
StyleGAN2 Generator + Discriminator & ${13.0 }$ \\ 
StyleGAN2 $\nabla_z D(G(z))$ (gradient for latent DOT) & ${18.5 }$ \\ 
StyleGAN2 Network $w^\varphi$ ($\star$) & ${1.5 }$ \\ 
StyleGAN2: $\nabla_z W(z)$ (gradient for latent GA on IW) ($\star$) & $2.7{ }$ \\ 
\hline
\end{tabular}
\end{center}
\caption{Inference time for one pass of different computational graphs. With the acceptance rate of the different methods, it allows to compute the runtime of these methods.}
\end{table}

\clearpage
\section{Qualitative results of latentGA.}
\begin{figure}[b]
    \centering
    \subfloat{\includegraphics[width=0.8\linewidth]{styleGan/gradient_descent_step_0_1__.jpeg}} \vspace{-0.2cm} \\
    \subfloat{\includegraphics[width=0.8\linewidth]{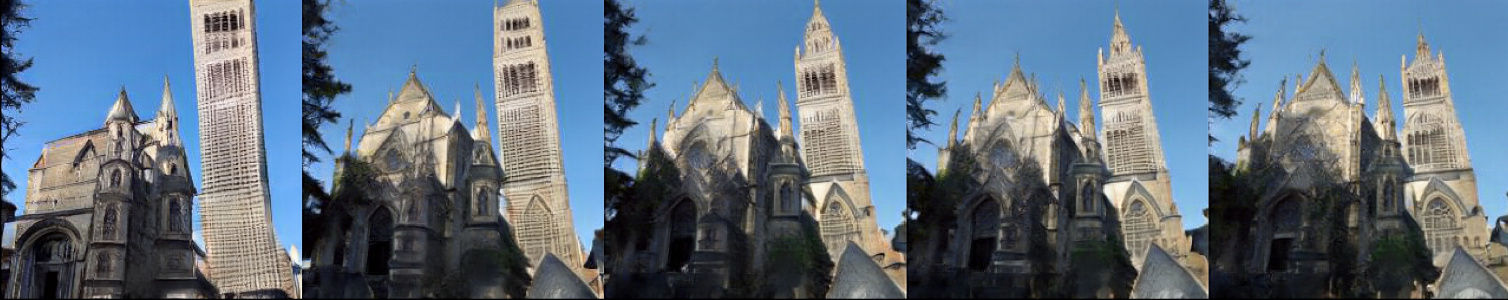}}  \vspace{-0.2cm}\\
    \subfloat{\includegraphics[width=0.8\linewidth]{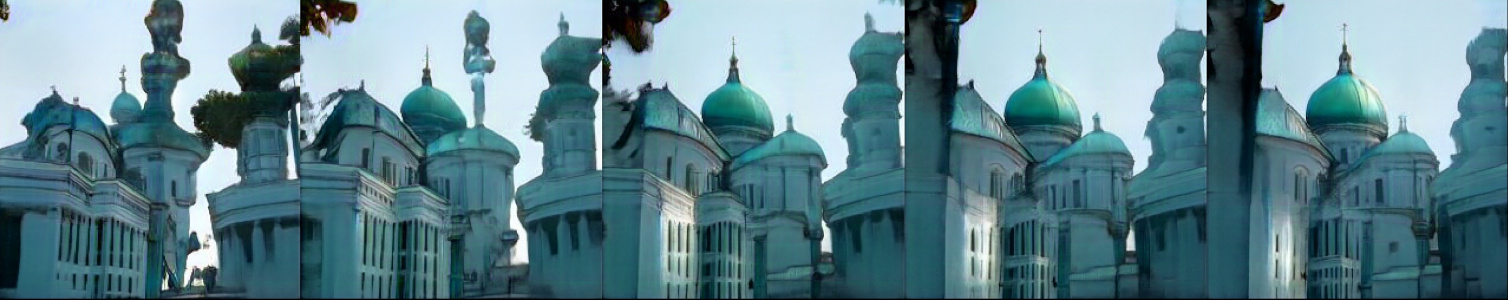}}  \vspace{-0.2cm}\\
    \subfloat{\includegraphics[width=0.8\linewidth]{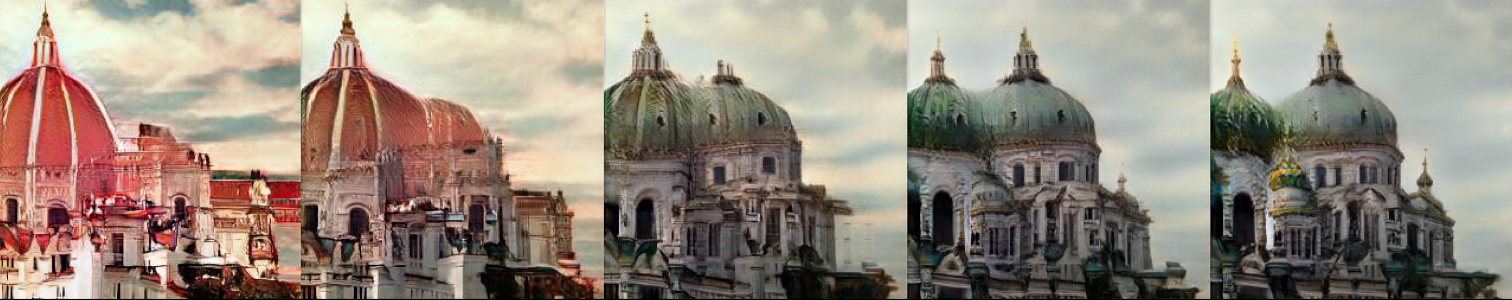}} \vspace{-0.2cm}\\
    \subfloat{\includegraphics[width=0.8\linewidth]{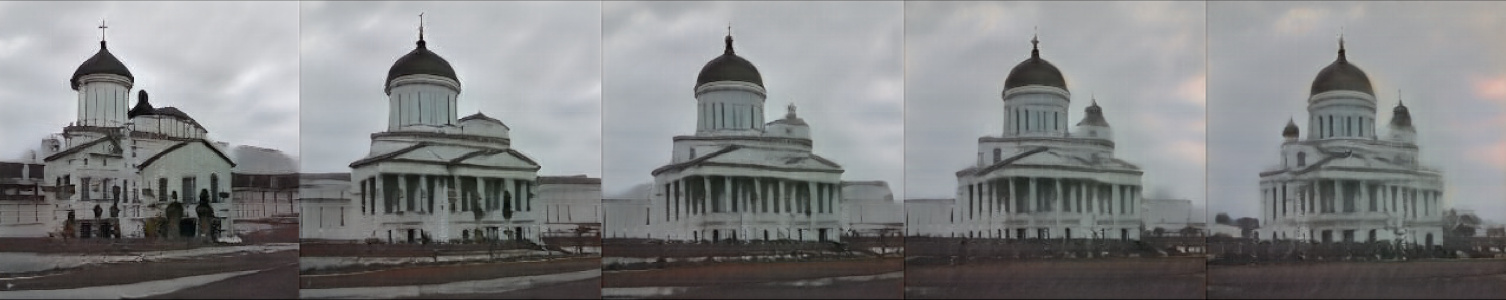}} \vspace{-0.2cm}\\
    \subfloat{\includegraphics[width=0.8\linewidth]{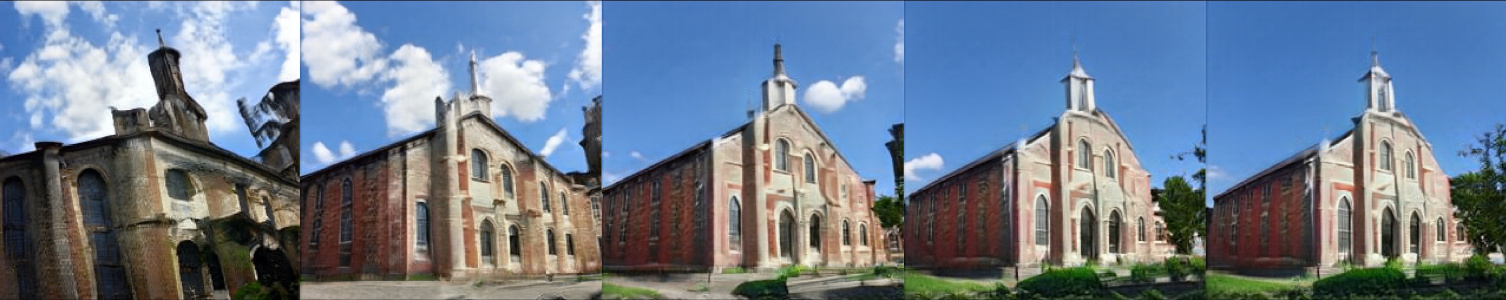}} \vspace{-0.2cm}\\
    \subfloat{\includegraphics[width=0.8\linewidth]{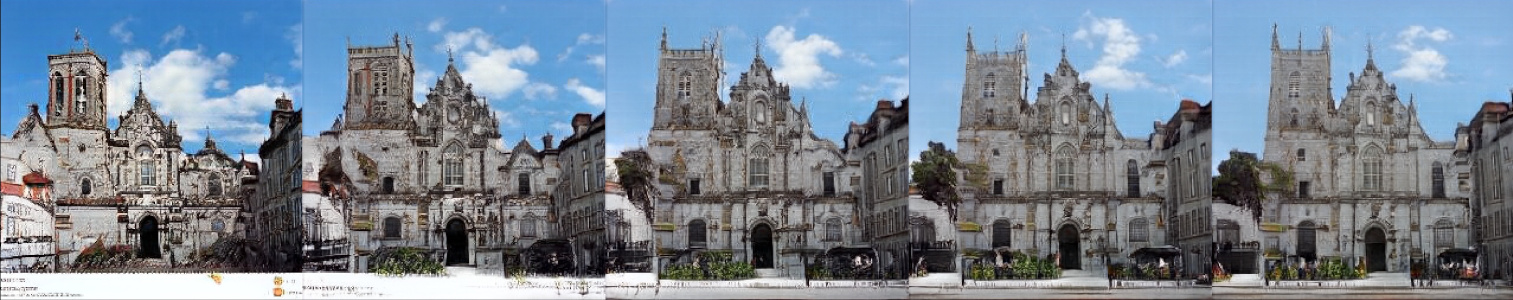}} \vspace{-0.2cm}\\
    \caption{Gradient ascent on latent importance weights (latentGA), on StyleGAN2 trained on LSUN Church.}
\end{figure}
\clearpage 
\newpage
\begin{figure}[b]
    \centering
    \subfloat{\includegraphics[width=0.8\linewidth]{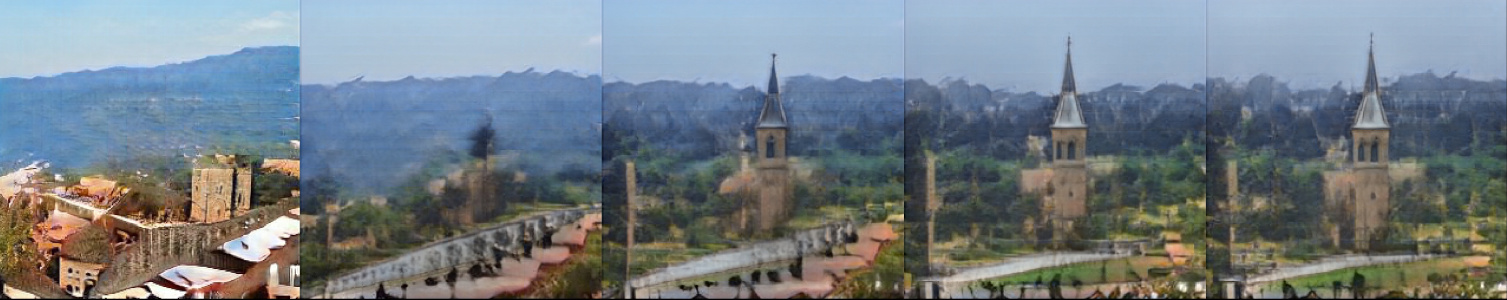}} \vspace{-0.2cm} \\
    \subfloat{\includegraphics[width=0.8\linewidth]{styleGan/gradient_descent_step_0_72_.jpeg}}  \vspace{-0.2cm}\\
    \subfloat{\includegraphics[width=0.8\linewidth]{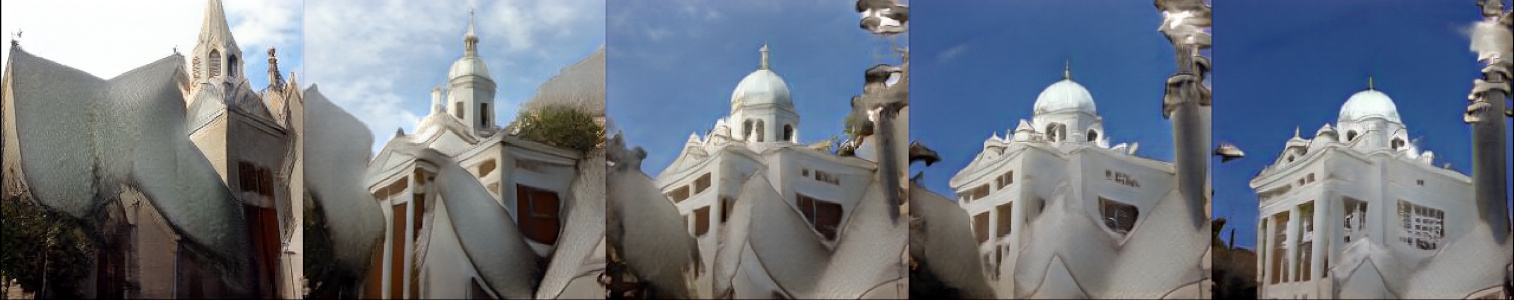}}  \vspace{-0.2cm}\\
    \subfloat{\includegraphics[width=0.8\linewidth]{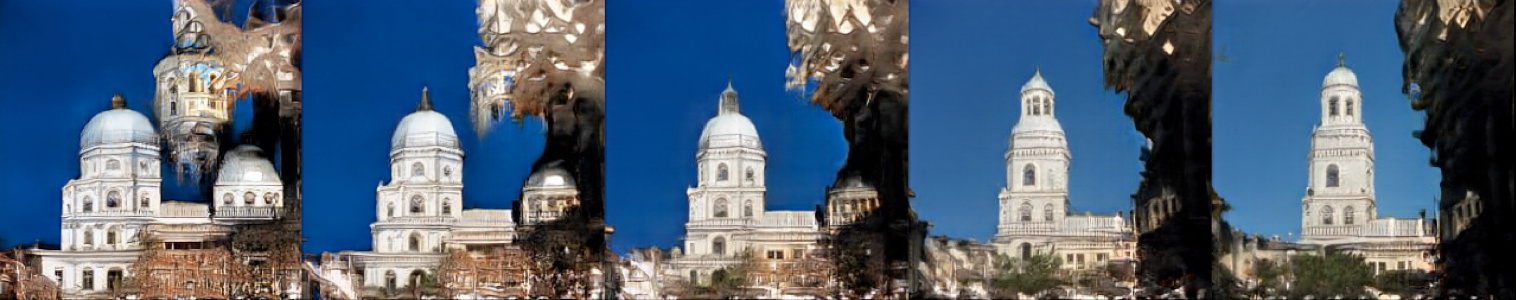}}
    \vspace{-0.2cm}\\ \subfloat{\includegraphics[width=0.8\linewidth]{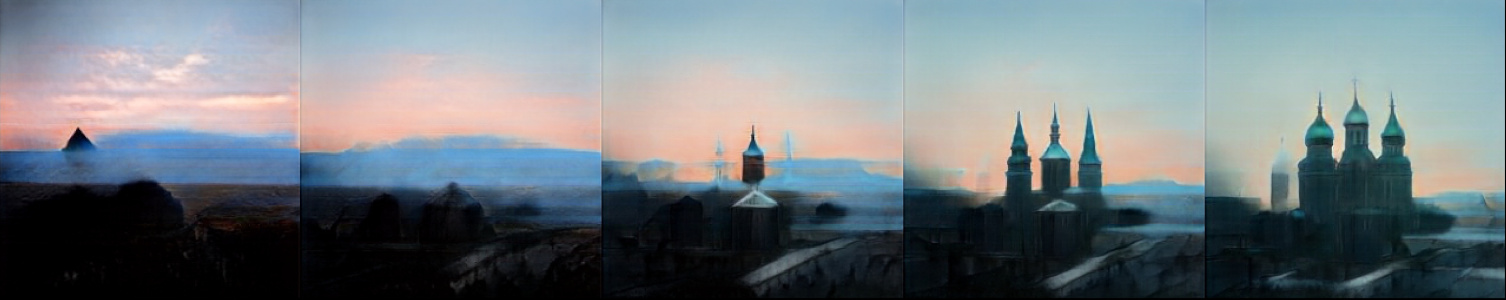}}
    \vspace{-0.2cm}\\
    \subfloat{\includegraphics[width=0.8\linewidth]{styleGan/gradient_descent_step_0_6_.jpeg}}
    \vspace{-0.2cm}\\
    \caption{Gradient ascent on latent importance weights (latentGA), on StyleGAN2 trained on LSUN Church.}
\end{figure}
\begin{figure}[b]
    \centering
    \subfloat[$P_a(z)=0.33$\hspace{1.2cm} $P_a(z)=0.40$\hspace{1.2cm}$P_a(z)=0.67$\hspace{1.2cm}$P_a(z)=0.75$\hspace{1.2cm}$P_a(z)= 0.87$]{\includegraphics[width=0.8\linewidth]{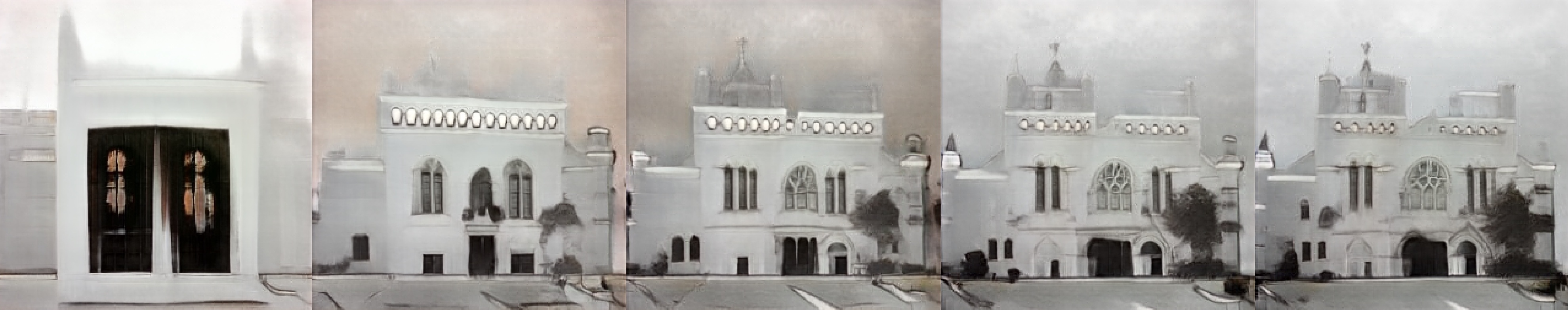}} \vspace{-0.2cm} \\
    \subfloat[$P_a(z)=0.09$\hspace{1.2cm} $P_a(z)=0.16$\hspace{1.2cm}$P_a(z)=0.31$\hspace{1.2cm}$P_a(z)=0.36$\hspace{1.2cm}$P_a(z)= 0.48$]{\includegraphics[width=0.8\linewidth]{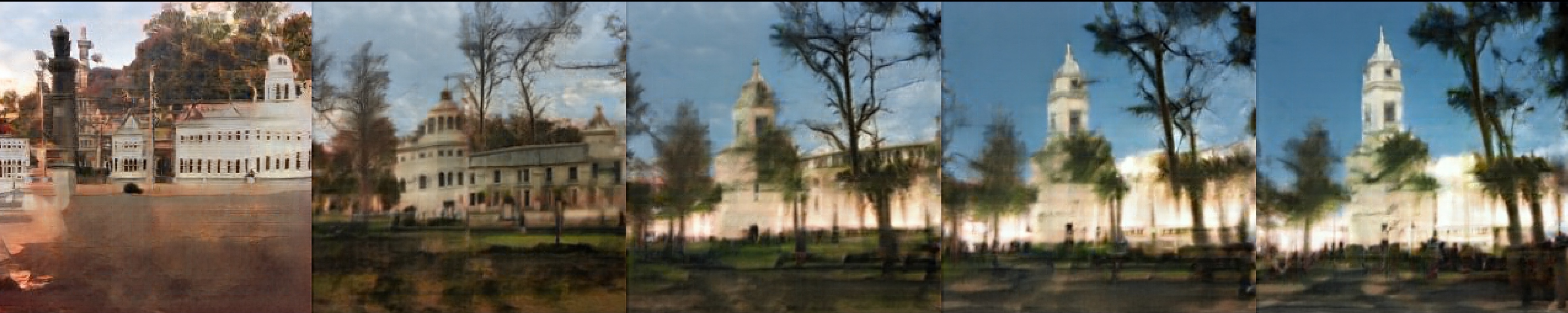}} \vspace{-0.2cm} \\
    \subfloat[$P_a(z)=0.51$\hspace{1.2cm} $P_a(z)=0.63$\hspace{1.2cm}$P_a(z)=0.70$\hspace{1.2cm}$P_a(z)=0.85$\hspace{1.2cm}$P_a(z)= 0.96$]{\includegraphics[width=0.8\linewidth]{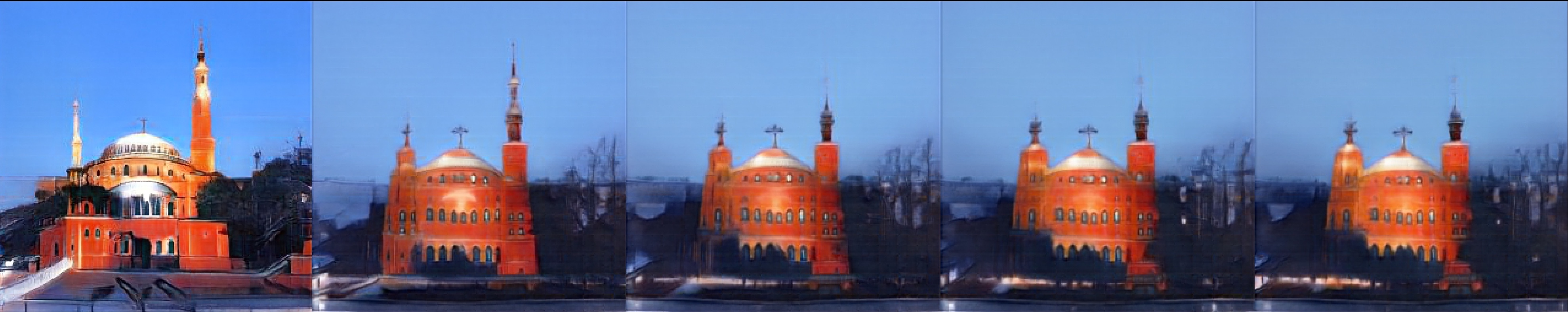}} \vspace{-0.2cm} \\
    \subfloat[$P_a(z)=0.41$\hspace{1.2cm} $P_a(z)=0.56$\hspace{1.2cm}$P_a(z)=0.59$\hspace{1.2cm}$P_a(z)=0.64$\hspace{1.2cm}$P_a(z)= 0.68$]{\includegraphics[width=0.8\linewidth]{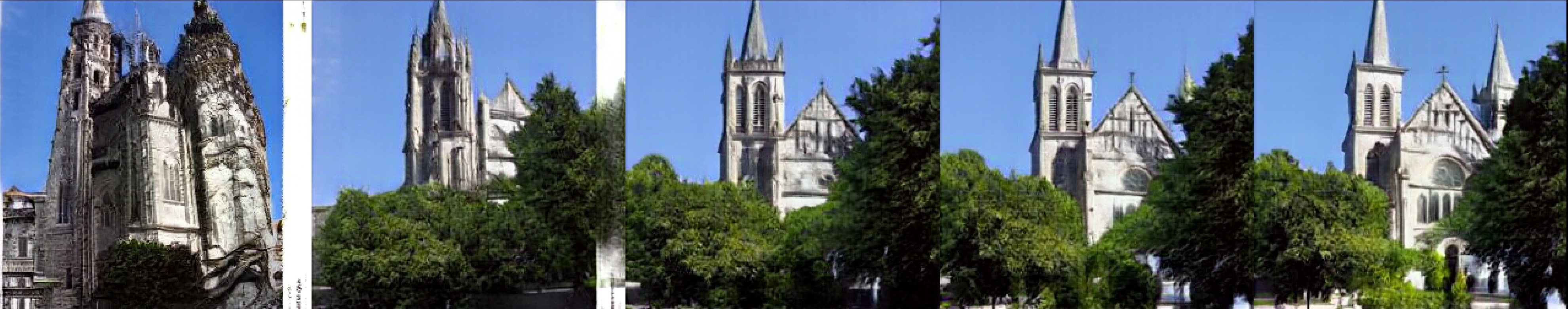}}
    \vspace{-0.2cm}\\
    \caption{Gradient ascent on latent importance weights (latentGA) on StyleGAN2 trained on LSUN Church. We visualize the evolution of the probability $P_a(z) = \frac{w^\varphi(z)}{m}$ associated to each sample.}
\end{figure}

\begin{figure}[]
    \centering
    \subfloat{\includegraphics[width=0.7\linewidth]{progGan/gradient_descent_step_30000_8.jpeg}} \\
    \subfloat{\includegraphics[width=0.7\linewidth]{progGan/gradient_descent_step_50000_11.jpeg}}\\
    \subfloat{\includegraphics[width=0.7\linewidth]{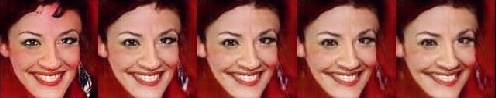}}\\
    \subfloat{\includegraphics[width=0.7\linewidth]{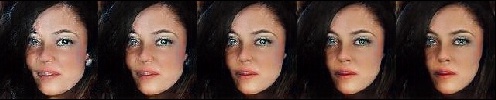}}\\
    \subfloat{\includegraphics[width=0.7\linewidth]{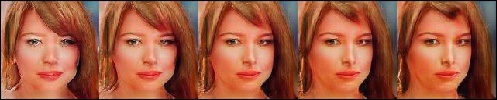}}\\
    \subfloat{\includegraphics[width=0.7\linewidth]{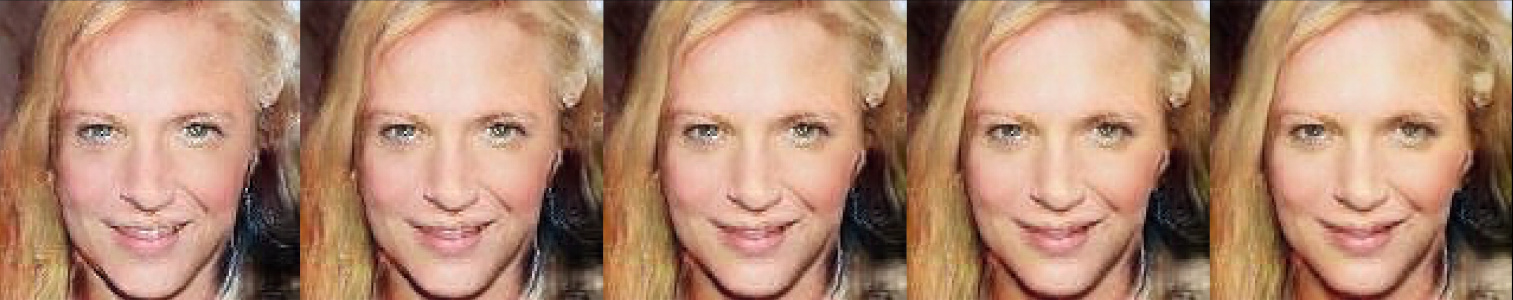}}\\
    \subfloat{\includegraphics[width=0.7\linewidth]{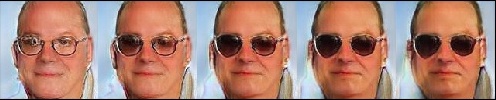}}\\
    \caption{Gradient ascent on latent importance weights (latentGA), on Progressive GAN trained on CelebA.}
\end{figure}
\clearpage
\newpage
\end{document}